\documentclass[10pt,twocolumn,letterpaper]{article}

\usepackage{cvpr}
\usepackage{times}
\usepackage{epsfig}
\usepackage{graphicx}
\usepackage{amsmath}
\usepackage{amssymb}

\usepackage[breaklinks=true,bookmarks=false]{hyperref}

\cvprfinalcopy 

\def\cvprPaperID{****} 

\begin{document}

\newcommand{\kun}[1]{\textcolor[rgb]{1,0,0}{#1}}
\newcommand{\zhe}[1]{\textcolor[rgb]{0,0,1}{#1}}

\newcommand\totalcharacter{1,018,402}
\newcommand\totalimage{32,285}
\newcommand\indicharacter{3,850}
\title{Chinese Text in the Wild}

\author{Tai-Ling Yuan\\
Tsinghua University\\
Beijing, China\\
{\tt\small yuantailing@gmail.com}
\and
Zhe Zhu\\
Tsinghua University\\
Beijing, China\\
{\tt\small ajex1988@gmail.com}
\and
Kun Xu\\
Tsinghua University\\
Beijing, China\\
{\tt\small xukun@tsinghua.edu.cn}
\and
Cheng-Jun Li\\
Tencent\\
Beijing, China\\
{\tt\small chengjunli@tencent.com}
\and
Shi-Min Hu\\
Tsinghua University\\
Beijing, China\\
{\tt\small shimin@tsinghua.edu.cn}
}

\maketitle

\begin{abstract}
   We introduce Chinese Text in the Wild, a very large dataset of Chinese text in street view images. While optical character recognition (OCR) in document images is well studied and many commercial tools are available, detection and recognition of text in natural images is still a challenging problem, especially for more complicated character sets such as Chinese text. Lack of training data has always been a problem, especially for deep learning methods which require massive training data.

   In this paper we provide details of a newly created dataset of Chinese text with about 1 million Chinese characters annotated by experts in over 30 thousand street view images.
   This is a challenging dataset with good diversity. It contains planar text, raised text, text in cities, text in rural areas, text under poor illumination, distant text, partially occluded text, etc.
   For each character in the dataset, the annotation includes its underlying character, its bounding box, and 6 attributes. The attributes indicate whether it has complex background, whether it is raised, whether it is handwritten or printed, etc.
   The large size and diversity of this dataset make it suitable for training robust neural networks for various tasks, particularly detection and recognition.  We give baseline results using several state-of-the-art networks, including  AlexNet, OverFeat, Google Inception and ResNet for character recognition, and YOLOv2 for character detection in images. Overall Google Inception has the best performance on recognition with 80.5\% top-1 accuracy, while YOLOv2 achieves an mAP of 71.0\% on detection. Dataset, source code and trained models will all be publicly available on the website\footnote{\href{https://ctwdataset.github.io/}{https://ctwdataset.github.io/}}.
\end{abstract}

\section{Introduction}
\begin{figure}[t!]
\centering
    \begin{tabular}{c c c c c c c c c}
    \includegraphics[width=0.04\textwidth]{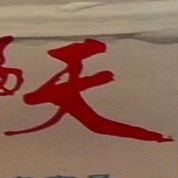} & \hspace{-0.12in} \includegraphics[width=0.04\textwidth]{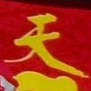} & \hspace{-0.12in} \includegraphics[width=0.04\textwidth]{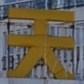} & \hspace{-0.12in} \includegraphics[width=0.04\textwidth]{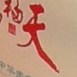} & \hspace{-0.12in} \includegraphics[width=0.04\textwidth]{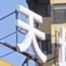} & \hspace{-0.12in} \includegraphics[width=0.04\textwidth]{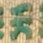} & \hspace{-0.12in} \includegraphics[width=0.04\textwidth]{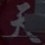} & \hspace{-0.12in} \includegraphics[width=0.04\textwidth]{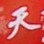} & \includegraphics[width=0.04\textwidth]{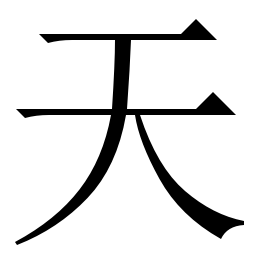}
    \\
    \includegraphics[width=0.04\textwidth]{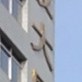} & \hspace{-0.12in} \includegraphics[width=0.04\textwidth]{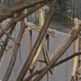} & \hspace{-0.12in} \includegraphics[width=0.04\textwidth]{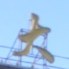} & \hspace{-0.12in} \includegraphics[width=0.04\textwidth]{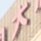} & \hspace{-0.12in} \includegraphics[width=0.04\textwidth]{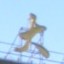} & \hspace{-0.12in} \includegraphics[width=0.04\textwidth]{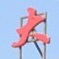} & \hspace{-0.12in} \includegraphics[width=0.04\textwidth]{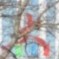} & \hspace{-0.12in} \includegraphics[width=0.04\textwidth]{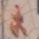} &
    \includegraphics[width=0.04\textwidth]{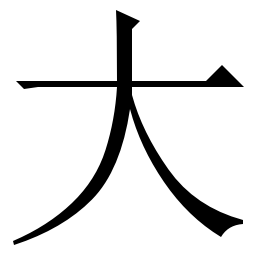} \\
    \includegraphics[width=0.04\textwidth]{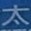} & \hspace{-0.12in} \includegraphics[width=0.04\textwidth]{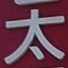} & \hspace{-0.12in} \includegraphics[width=0.04\textwidth]{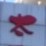} & \hspace{-0.12in} \includegraphics[width=0.04\textwidth]{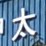} & \hspace{-0.12in} \includegraphics[width=0.04\textwidth]{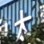} & \hspace{-0.12in} \includegraphics[width=0.04\textwidth]{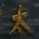} & \hspace{-0.12in} \includegraphics[width=0.04\textwidth]{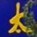} &\hspace{-0.12in} \includegraphics[width=0.04\textwidth]{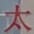} &
    \includegraphics[width=0.04\textwidth]{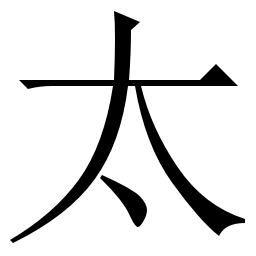}
    \end{tabular}
    \caption{High intra-class variance versus low inter-class variance. Each row shows instances of a Chinese character. The first character differs from the second character by a single stroke, and the second character differs from the third character by another stroke. While the three characters are very similar in shape, the instances of the same character have very different appearance, due to color, font,
    occlusion, and background differences, etc. The most right column shows the corresponding Chinese character.
    }
    \label{fig:fine_grain}
\end{figure}

Automatic text detection and recognition is an important task in computer vision, with many uses ranging from autonomous driving to book digitization. This problem has been extensively studied, and has been  divided into two problems at different levels of difficulty:  text detection and recognition in document images, and text detection and recognition in natural images. The former is less challenging and many commercial tools are already available. However, text detection and recognition in natural images are still challenging.
For example, a character may have very different appearances in different images due to style, font, resolution, or illumination differences; characters may also be partially occluded, distorted, or have complex background, which makes detection and recognition even harder.
Sometimes we even have to deal with  high intra-class versus low inter-class differences~\cite{DBLP:journals/corr/CuiZLB15}. As shown in Figure~\ref{fig:fine_grain}, the three characters differ a little, but the instances of the same character could have large appearance differences.


The past few years have witnessed a boom of deep learning in many fields, including image classification, speech recognition, and so on.
Very deep networks with tens or even more than a hundred layers (such as VGG-19, Google Inception or ResNet) have nice modeling capacity, and have shown promising performance in a variety of detection, classification, recognition tasks.
These models need massive amount of data for training. Availability of data is, indeed, a key factor in the success of deep neural networks. The public datasets, such as Image-Net dataset~\cite{imagenet_cvpr09}, the Microsoft COCO dataset~\cite{DBLP:journals/corr/LinMBHPRDZ14} and the ADE20K dataset~\cite{zhou2016semantic}, have become a key driver for progress in computer vision.

In this paper we present a large dataset of Chinese text in natural images, named \emph{Chinese Text in the Wild} (CTW).
The dataset contains \totalimage\ images with \totalcharacter\ Chinese characters, going much beyond previous datasets.
The images are from Tencent Street View. They are captured from tens of different cities in China, without preference for any particular purpose.
The dataset is a challenging dataset, due to its diversity and complexity. It contains planar text, raised text, text in cities, text in rural areas, text under poor illumination, distant text, partially occluded text, etc.
For each image, we annotate all Chinese texts in it. For each Chinese character, we annotate its underlying character, its bounding box, and 6 attributes to indicate whether it is occluded, having complex background, distorted, 3D raised, wordart, and handwritten, respectively.

We have used the dataset as a basis to train deep models using several state-of-the-art approaches for character recognition, and character detection in images. These models are also presented as  baseline algorithms. The dataset, source code and baseline algorithms will all be publicly available. We expect the dataset to greatly stimulate future development of
detection and recognition algorithms of Chinese texts in natural images.

The rest of this paper is organized as follows. We discuss related work in Section~\ref{sec:related_work}, and give details of our dataset in Section~\ref{sec:dataset}.
The baseline algorithms trained using our dataset and experimental results are given in Section~\ref{sec:experiment},
and conclusions are presented in Section~\ref{sec:discussion}.

\section{Related work}
\label{sec:related_work}
Text detection and recognition has received much attention during the past decades in the computer vision community, albeit mostly for English text and numbers. We briefly review both benchmark datasets and approaches from recent years. Here, we treat text recognition in documents as a separate well studied problem, and only focus on text detection and recognition in natural images.

\subsection{Datasets of text in natural images}


Datasets of text in natural images could be classified into two categories: those that only contain real world text~\cite{Smith2016,veit2016cocotext,MishraBMVC12,Lucas2005}, and those that contain synthetic text~\cite{de2009character,Jaderberg14c}. Images in these datasets are mainly of two kinds: Internet images, and Google Street View images. For example, the SVHN~\cite{37648} and SVT~\cite{6126402} datasets  utilize Google Street View images, augmented by  annotations for numbers and text, respectively. Most previous datasets target English text in the Roman alphabet, and digits, although several recent datasets consider text in other languages and character sets. Notable amongst them are the KAIST scene text dataset~\cite{jung2011touch} for Korean text, FSNS~\cite{Smith2016} for French text, and MSRA-TD500~\cite{Yao:2012:DTA:2354409.2354851} for Chinese text. However, MSRA-TD500 dataset~\cite{Yao:2012:DTA:2354409.2354851} only contains 500 natural images, which is far from sufficient for training deep models such as convolutional neural networks. In contrast, our dataset contains over 30 thousand images and about 1 million Chinese characters.

\subsection{Text detection and recognition}
Text detection and recognition approaches can be classified as those approaches that use hand-crafted features, and those approaches that use automatically learned features (as deep learning does). We draw a distinction between \emph{text detection}, detecting a region of an image that (potentially) contains text, and \emph{text recognition}, determining which characters and text are present, typically using the cropped areas returned by text detection.

The most widely used approach to text detection based on hand-crafted features is the \emph{stroke width transform} (SWT)~\cite{5540041}. The SWT transforms an image into a new stroke-width image with equal size, in which the value of each pixel is the stroke width associated with the original pixel. This approach works quite well for relatively clean images containing English characters and digits, but often fails on more cluttered images. Another widely used approach is to seek text as maximally stable extremal regions (MSERs)~\cite{Matas2004761,6116200,6471224,Neumann2011}. Such MSERs always contain non-text regions, so a robust filter is needed for candidate text region selection. Recently, deep learning based approaches have been adopted for text detection, including both fully convolutional networks (FCN)~\cite{zhang2016multi} and cascaded convolutional text networks (CCTN)~\cite{DBLP:journals/corr/HeH0Y16}.

Given cropped text, recognition methods for general objects can be adapted to text recognition. Characters and words are at two different levels in English, and different approaches have been proposed for character recognition and word recognition separately. For character recognition, both SVM based approaches~\cite{http://dx.doi.org/10.5244/C.26.13} and part-based models~\cite{6619225} have been applied and found to work well. Word recognition provides additional contextual information, so Bayesian inferencing~\cite{1211512}, conditional random fields~\cite{6247990} and graph models~\cite{Lee:2008:CCH:1332129.1332191} can now be used.

A recent trend is to focus on `end-to-end' recognition~\cite{6460871}; more can be found in a detailed survey by Ye and Doermann~\cite{10.1109/TPAMI.2014.2366765}.

\section{Chinese Text in the Wild Dataset}
\label{sec:dataset}

In this section, we present Chinese Text in the Wild (CTW), a very large dataset of Chinese text in street view images.
We will discuss how the images are selected, annotated, split into training and testing sets, and we also provide statistics of the dataset.
For denotation clearness, we refer to each unique Chinese character as a \emph{character category} or as a \emph{category}, and refer to an observed instance of a Chinese character in an image as a \emph{character instance}, or as an \emph{instance}.


\subsection{Image selection}
\label{selection}

We have collected 122,903 street view images from Tencent Street View. Among them, 98,903 images are from the Tsinghua-Tencent 100K dataset~\cite{Zhu_2016_CVPR}, and 24,000 directly from Tencent Street View. These images are captured from tens of different cities in China, and each image has a resolution of  $2048 \times 2048$. We manually check all street view images, and remove those images which do not contain any Chinese characters.
Besides, since the street view images were captured at fixed intervals (i.e., 10 to 20 meters), successive images may have large duplicated areas. Hence, we manually check each pair of successive images, if duplicated areas cover more than 70\% of the total image size, we also remove one image.
Finally, \totalimage\ images are selected.

\subsection{Annotation}

\begin{figure*}
    \centering
    \includegraphics[width=1.0\textwidth]{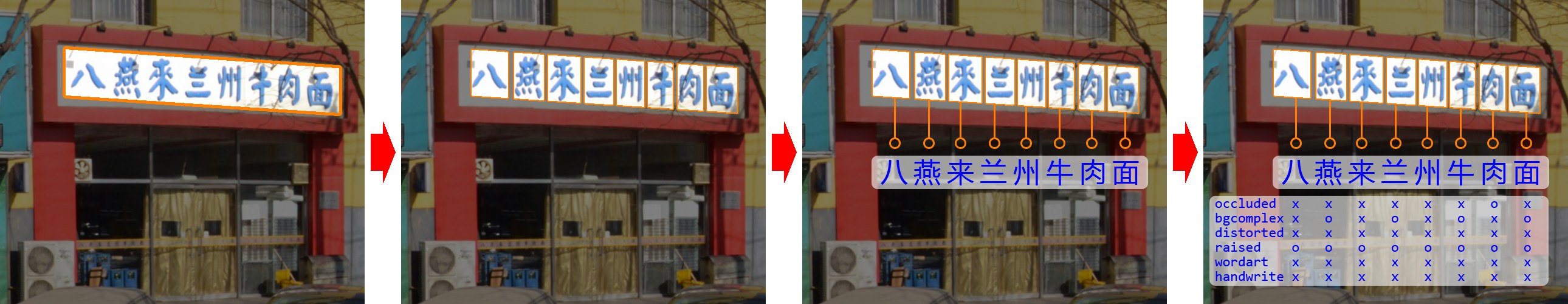} \\
    (a)~~~~~~~~~~~~~~~~~~~~~~~~~~~~~~~~~~~~~~~~~~~~~~~~(b)~~~~~~~~~~~~~~~~~~~~~~~~~~~~~~~~~~~~~~~~~~~~~~~~(c)~~~~~~~~~~~~~~~~~~~~~~~~~~~~~~~~~~~~~~~~~~~~~~~~(d)
    \caption{Annotation pipeline: drawing a bounding box for the sentence (a), drawing a bounding box for each character instance (b), labeling its corresponding character category (c), and labeling its attributes (d).}
    \label{fig:pipeline}
\end{figure*}

We now describe the annotation process in detail. For each image, all Chinese character instances are annotated.
Characters in English and other languages are not annotated. Our annotation pipeline is illustrated in Figure~\ref{fig:pipeline}. A bounding box is first drawn around a sentence of Chinese text. Next, for each character instance, a more tight bounding box is drawn around it, and its corresponding character instance and its attributes are also specified.

\begin{figure}
\centering
\begin{tabular}{c c c c c c c c c c}
\includegraphics[width=.045\textwidth]{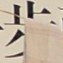} &  \hspace{-0.2in}
\includegraphics[width=.045\textwidth]{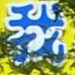} &  \hspace{-0.2in}
\includegraphics[width=.045\textwidth]{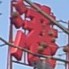} &  \hspace{-0.2in}
\includegraphics[width=.045\textwidth]{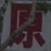} &  \hspace{-0.2in}
\includegraphics[width=.045\textwidth]{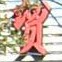} &  \hspace{-0.1in}
\includegraphics[width=.045\textwidth]{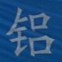} &  \hspace{-0.2in}
\includegraphics[width=.045\textwidth]{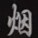} &  \hspace{-0.2in}
\includegraphics[width=.045\textwidth]{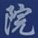} &  \hspace{-0.2in}
\includegraphics[width=.045\textwidth]{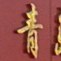} &  \hspace{-0.2in}
\includegraphics[width=.045\textwidth]{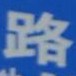} \\
\multicolumn{5}{c}{(a) occluded} & \multicolumn{5}{c}{(b) not occluded} \\
\includegraphics[width=.045\textwidth]{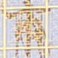} &  \hspace{-0.2in}
\includegraphics[width=.045\textwidth]{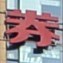} &  \hspace{-0.2in}
\includegraphics[width=.045\textwidth]{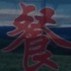} &  \hspace{-0.2in}
\includegraphics[width=.045\textwidth]{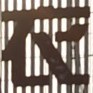} &  \hspace{-0.2in}
\includegraphics[width=.045\textwidth]{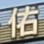} &  \hspace{-0.1in}
\includegraphics[width=.045\textwidth]{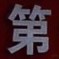} &  \hspace{-0.2in}
\includegraphics[width=.045\textwidth]{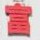} &  \hspace{-0.2in}
\includegraphics[width=.045\textwidth]{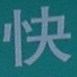} &  \hspace{-0.2in}
\includegraphics[width=.045\textwidth]{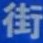} &  \hspace{-0.2in}
\includegraphics[width=.045\textwidth]{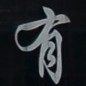} \\
\multicolumn{5}{c}{(c) complex background} & \multicolumn{5}{c}{(d) clean background} \\
\includegraphics[width=.045\textwidth]{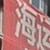} &  \hspace{-0.2in}
\includegraphics[width=.045\textwidth]{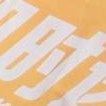} &  \hspace{-0.2in}
\includegraphics[width=.045\textwidth]{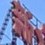} &  \hspace{-0.2in}
\includegraphics[width=.045\textwidth]{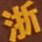} &  \hspace{-0.2in}
\includegraphics[width=.045\textwidth]{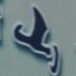} &  \hspace{-0.1in}
\includegraphics[width=.045\textwidth]{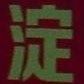} &  \hspace{-0.2in}
\includegraphics[width=.045\textwidth]{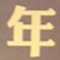} &  \hspace{-0.2in}
\includegraphics[width=.045\textwidth]{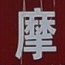} &  \hspace{-0.2in}
\includegraphics[width=.045\textwidth]{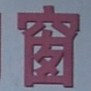} &  \hspace{-0.2in}
\includegraphics[width=.045\textwidth]{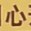} \\
\multicolumn{5}{c}{(e) distorted} & \multicolumn{5}{c}{(f) frontal} \\
\includegraphics[width=.045\textwidth]{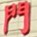} &  \hspace{-0.2in}
\includegraphics[width=.045\textwidth]{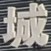} &  \hspace{-0.2in}
\includegraphics[width=.045\textwidth]{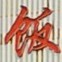} &  \hspace{-0.2in}
\includegraphics[width=.045\textwidth]{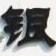} &  \hspace{-0.2in}
\includegraphics[width=.045\textwidth]{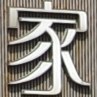} &  \hspace{-0.1in}
\includegraphics[width=.045\textwidth]{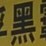} &  \hspace{-0.2in}
\includegraphics[width=.045\textwidth]{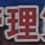} &  \hspace{-0.2in}
\includegraphics[width=.045\textwidth]{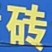} &  \hspace{-0.2in}
\includegraphics[width=.045\textwidth]{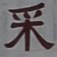} &  \hspace{-0.2in}
\includegraphics[width=.045\textwidth]{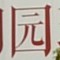} \\
\multicolumn{5}{c}{(g) 3D raised} & \multicolumn{5}{c}{(h) planar} \\
\includegraphics[width=.045\textwidth]{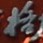} &  \hspace{-0.2in}
\includegraphics[width=.045\textwidth]{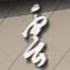} &  \hspace{-0.2in}
\includegraphics[width=.045\textwidth]{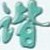} &  \hspace{-0.2in}
\includegraphics[width=.045\textwidth]{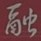} &  \hspace{-0.2in}
\includegraphics[width=.045\textwidth]{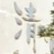} &  \hspace{-0.1in}
\includegraphics[width=.045\textwidth]{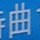} &  \hspace{-0.2in}
\includegraphics[width=.045\textwidth]{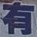} &  \hspace{-0.2in}
\includegraphics[width=.045\textwidth]{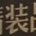} &  \hspace{-0.2in}
\includegraphics[width=.045\textwidth]{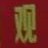} &  \hspace{-0.2in}
\includegraphics[width=.045\textwidth]{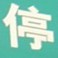} \\
\multicolumn{5}{c}{(i) wordart} & \multicolumn{5}{c}{(j) not wordart} \\
\includegraphics[width=.045\textwidth]{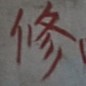} &  \hspace{-0.2in}
\includegraphics[width=.045\textwidth]{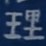} &  \hspace{-0.2in}
\includegraphics[width=.045\textwidth]{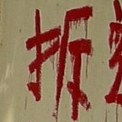} &  \hspace{-0.2in}
\includegraphics[width=.045\textwidth]{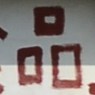} &  \hspace{-0.2in}
\includegraphics[width=.045\textwidth]{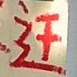} &  \hspace{-0.1in}
\includegraphics[width=.045\textwidth]{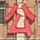} &  \hspace{-0.2in}
\includegraphics[width=.045\textwidth]{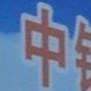} &  \hspace{-0.2in}
\includegraphics[width=.045\textwidth]{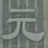} &  \hspace{-0.2in}
\includegraphics[width=.045\textwidth]{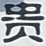} &  \hspace{-0.2in}
\includegraphics[width=.045\textwidth]{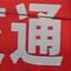} \\
\multicolumn{5}{c}{(k) handwritten} & \multicolumn{5}{c}{(l) printed} \\
\end{tabular}
\caption{Examples with different attributes.
}
\label{fig:attributes}
\end{figure}

There are six attributes to annotate, which are occlusion attribute, complex background attribute, distortion attribute, raised attribute, wordart attribute, and handwritten attribute.
For each character, \emph{yes} or \emph{no} is specified for each attribute. The occlusion attribute indicates whether the character is occluded, partially occluded by other objects or not. The complex background attribute indicates whether the character has complex background, shadows on it or not. The distortion attribute indicates whether the character is distorted, rotated or it is frontal. The raised attribute indicates whether the character is 3D raised or it is planar. The wordart attribute indicates whether the character uses a artistic style or uses a traditional font. The handwritten attribute indicates whether the character is handwritten or printed. Character examples of each attribute are illustrated in Figure~\ref{fig:attributes}.
We provide these attributes since the texts have large appearance variations due to color, font, occlusion, and background differences, etc. With the help of these attributes, it will be easier to analyze the algorithm performance on different styles of texts. Researcher may also design algorithms for specific styles of Chinese texts, i.e., 3D raised texts.


\begin{figure}[t!]
    \centering
    \includegraphics[width=.5\textwidth]{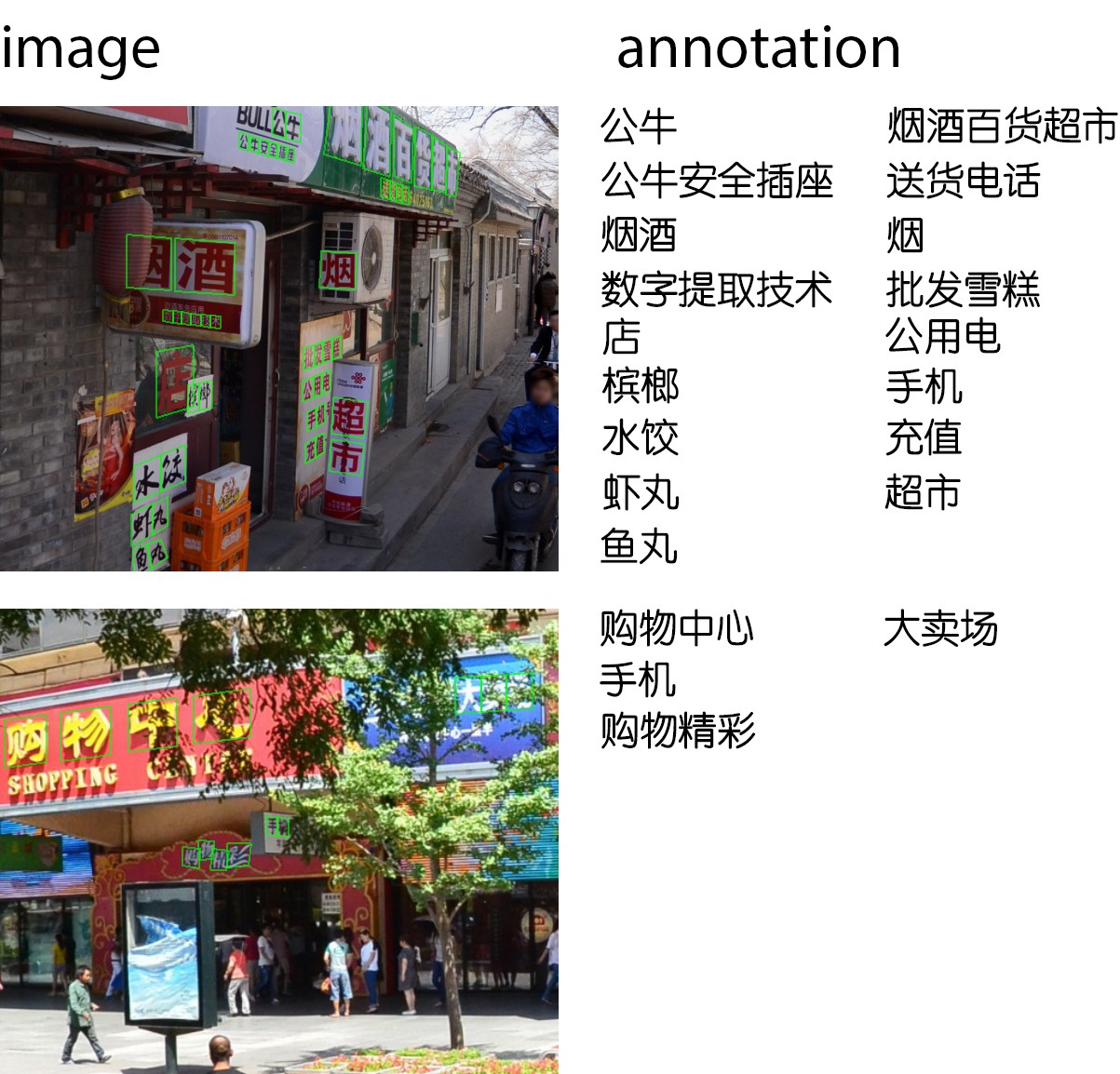}
    \caption{Left: two images in our dataset. Right: corresponding ground truth annotation.}
    \label{fig:teaser}
\end{figure}

In order to ensure high quality, we invite 40 annotation experts for the annotation process. They are employed by a professional image annotation company and are well trained for image annotations tasks.
We also invite two inspectors to verify the quality of annotations.
Before annotating, we first invite them to take a training session on annotation instructions.
The whole annotation process took about 2 months. In total, \totalcharacter\ Chinese character instances are annotated.
Figure~\ref{fig:teaser} shows two images in our dataset and the corresponding annotation.

\subsection{Dataset splitting} \label{subsec:data_split}

We split our dataset to a training set and a testing set. The testing set is further split into a recognition testing set for the recognition task (Section~\ref{subsec:recognition}) and a detection testing set for the detection task (Section~\ref{subsec:end_to_end_det_and_rec}).  We set the ratio of the sizes of the three sets to $8:1:1$. We randomly distribute all the images into the three sets according to the ratio.
To avoid correlation between training and testing images, we constrain that the images captured on the same street must be in the same set. After splitting, the training set contains 25,887 images with 812,872 Chinese characters, the recognition testing set contains 3,269 images with 103,519 Chinese characters, and the detection testing set contains 3,129 images with 102,011 Chinese characters.

\subsection{Statistics}

Our CTW dataset contains \totalimage\ images with \totalcharacter\ Chinese character instances. It contains \indicharacter\ character categories (i.e., unique Chinese characters).

\begin{figure*}
    \centering
    \includegraphics[width=0.9\textwidth]{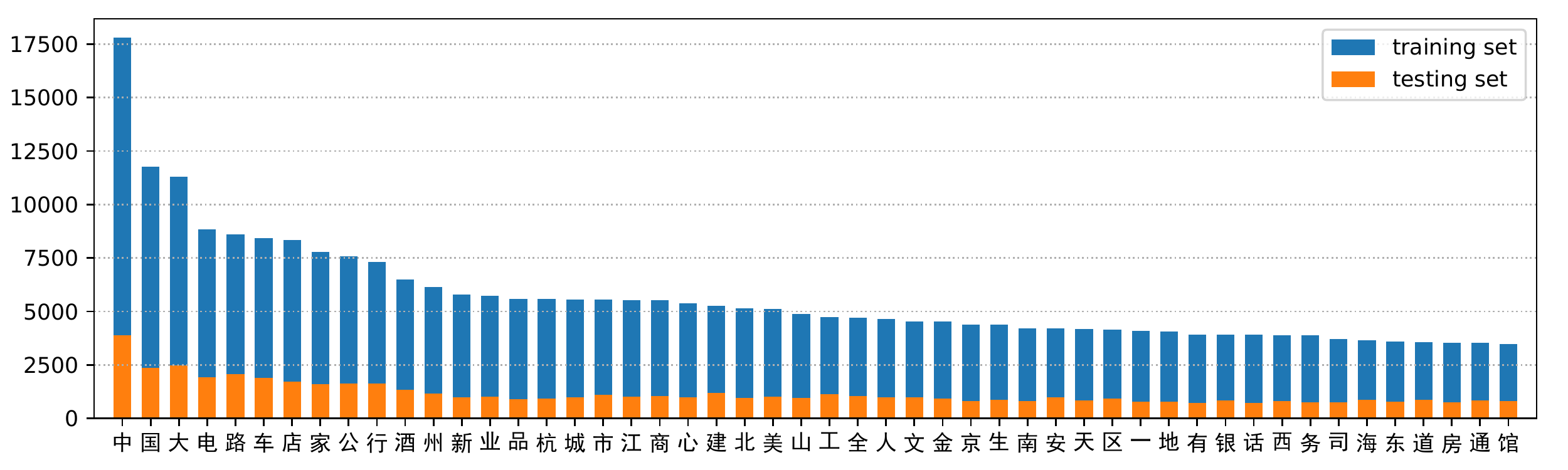}
    \caption{Number of character instances for the 50 most frequent observed character categories in our dataset.}
    \label{fig:instance_per_category}
\end{figure*}

\begin{figure}[t!]
\centering
\begin{tabular}{c}
\includegraphics[width=.45\textwidth]{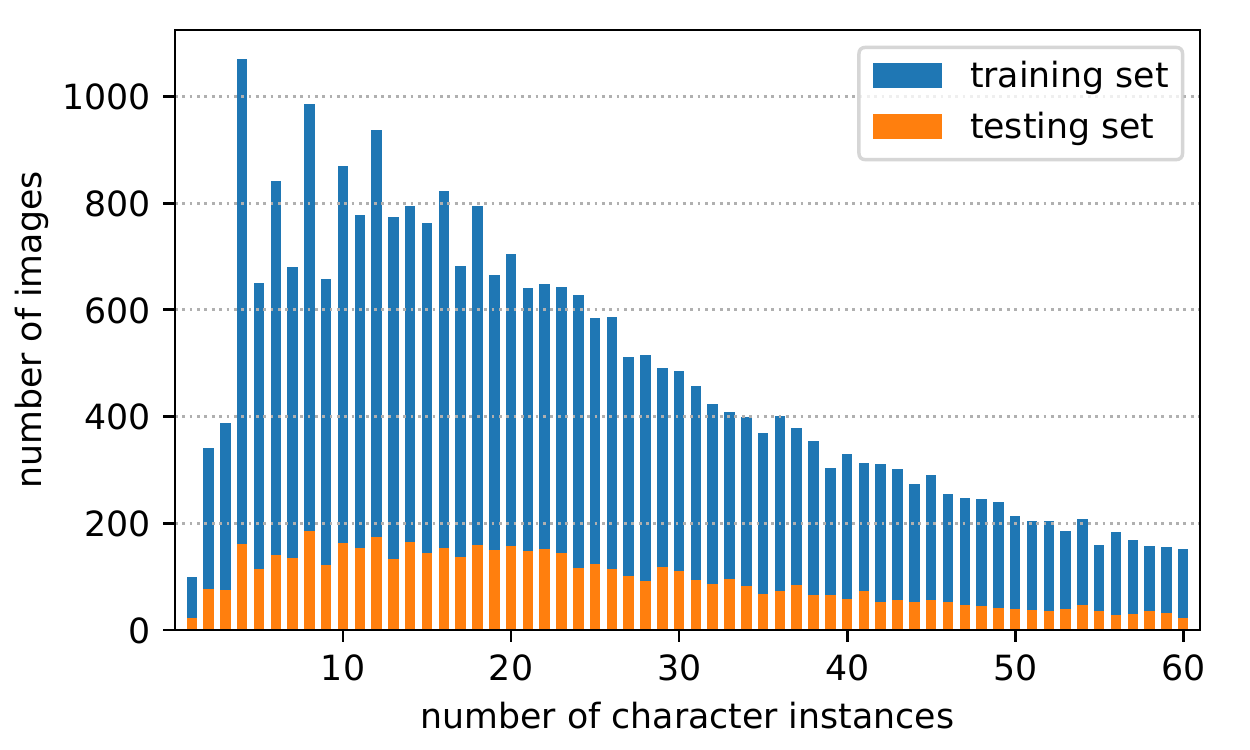} \\
    (a) \\
\includegraphics[width=.45\textwidth]{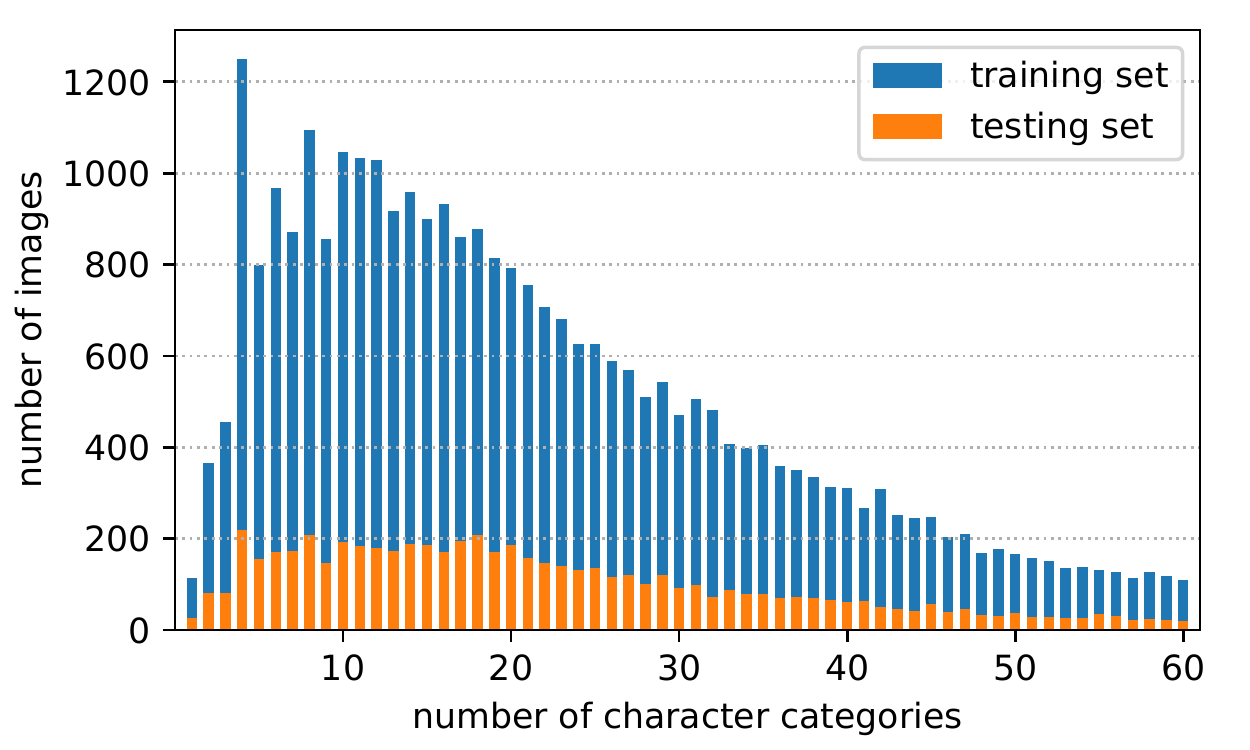} \\
    (b)
\end{tabular}
\caption{Histograms. (a) Number of images containing specific number of character instances; (b) Number of images containing specific number of character categories.}
\label{fig:instances_categories}
\end{figure}

\begin{figure}[t!]
    \centering
    \includegraphics[width=.5\textwidth]{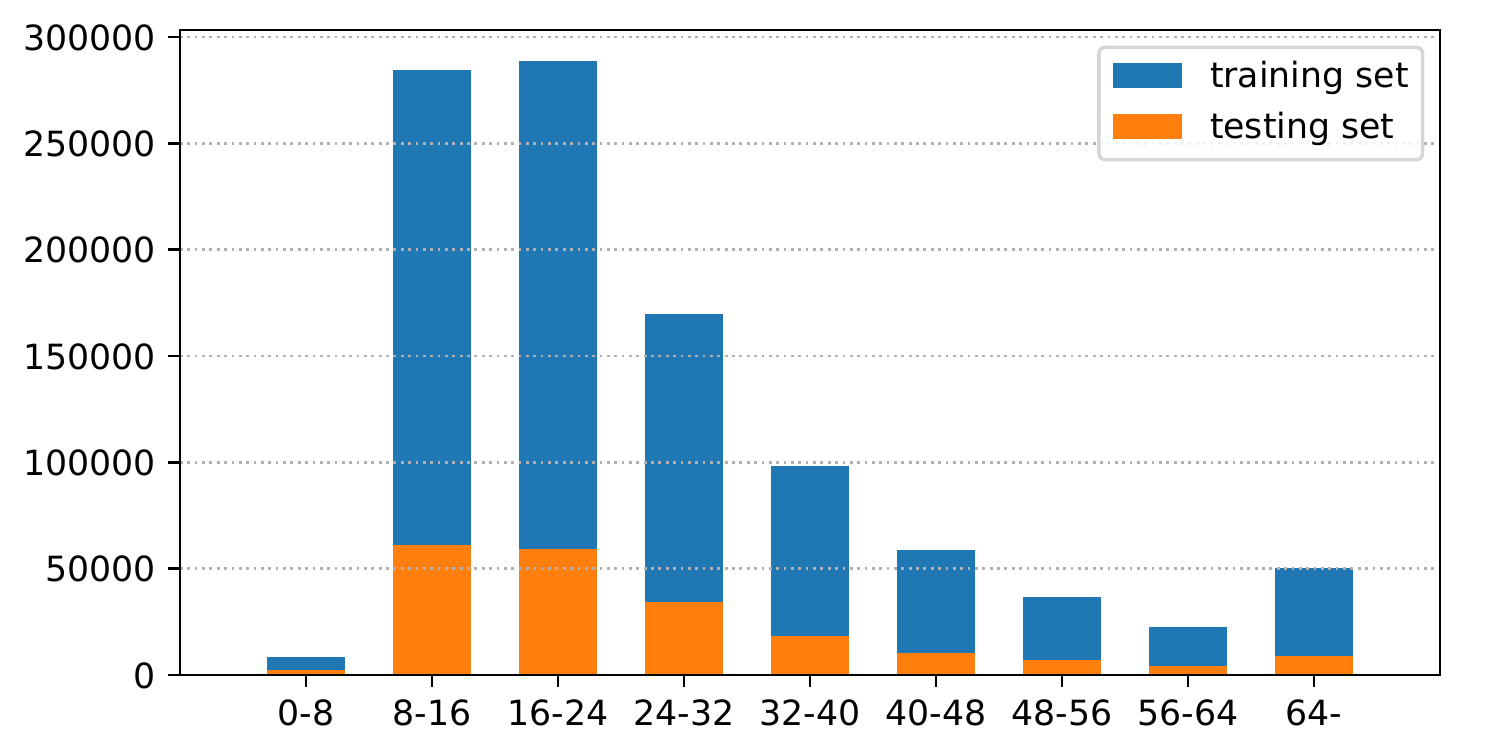}
    \caption{The number of character instances with different sizes. The size is measured by the long side of its bounding box in pixels.}
    \label{fig:instance_by_size}
\end{figure}

\begin{figure}[t!]
    \centering
    \includegraphics[width=.5\textwidth]{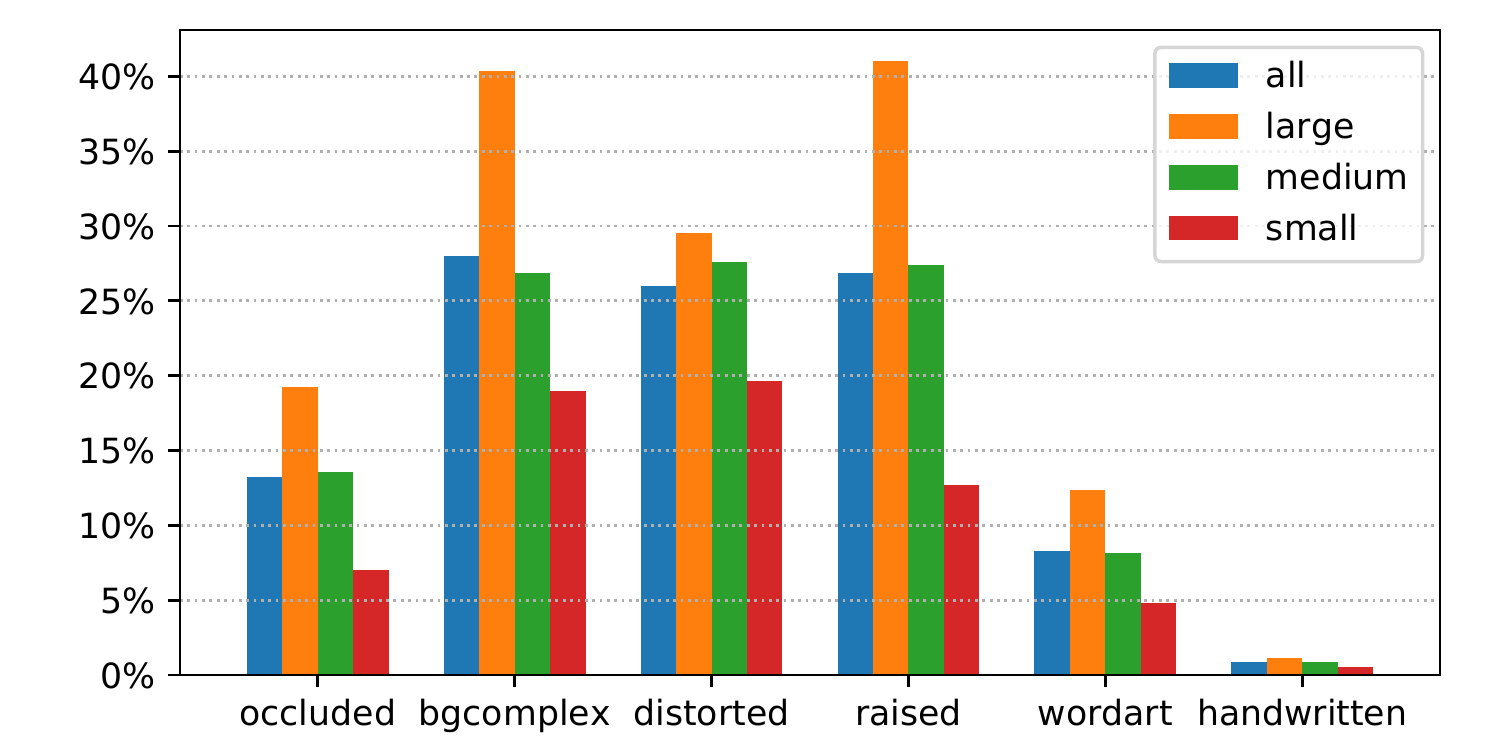}
    \caption{The percentage of character instances with different attributes in all/large/medium/small character instances, respectively. Small, medium, and large refer to character size $<16$, $\in [16,32)$ and $\ge32$, respectively.
    }
    \label{fig:instance_by_attributes}
\end{figure}

In Figure~\ref{fig:instance_per_category}, for the top 50 most frequent observed character categories, we show the number of character instances in each category in the training set and in the testing set, respectively. 
In Figure~\ref{fig:instances_categories} (a), we show the number of images contains specific number of character instances in the training set and in the testing set, respectively.
In Figure~\ref{fig:instances_categories} (b), we show the number of images containing specific number of character categories in the training set and in the testing set, respectively.
In Figure~\ref{fig:instance_by_size}, we provide the number of character instances with different sizes in the training set and in the testing set, respectively, where the size is measured by the long side of its bounding box in pixels.
In Figure~\ref{fig:instance_by_attributes}, we provide the percentage of character instances with different attributes in all/large/medium/small character instances, respectively. Small, medium, and large refer to character size $<16$, $\in [16,32)$ and $\ge32$, respectively. We could find that large character instances are more likely to have complex attributes. For example, in all character instances, 13.2\% of them are occluded, while in all large character instances, a higher proportion (19.2\%) of them are occluded.



\begin{figure}
\centering
\begin{tabular}{c c c c c }
\hspace{-0.1in} \rotatebox{90}{~~~~~~~(a)} & \hspace{-0.17in} \includegraphics[width=.11\textwidth]{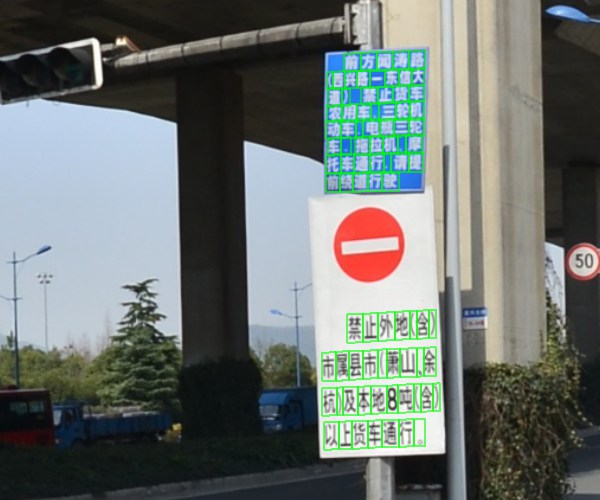} &  \hspace{-0.17in}
\includegraphics[width=.11\textwidth]{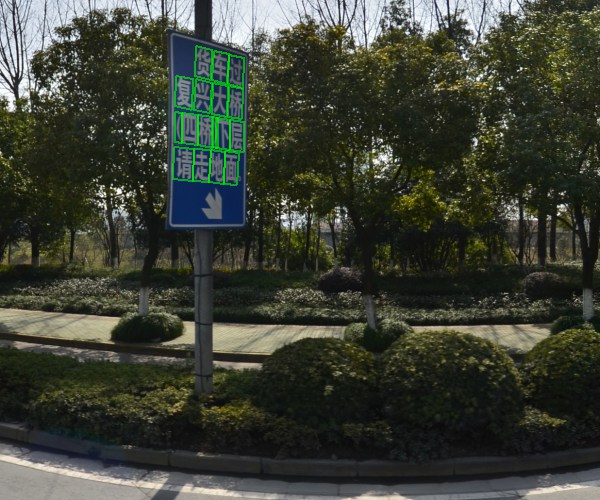} & \hspace{-0.17in}
\includegraphics[width=.11\textwidth]{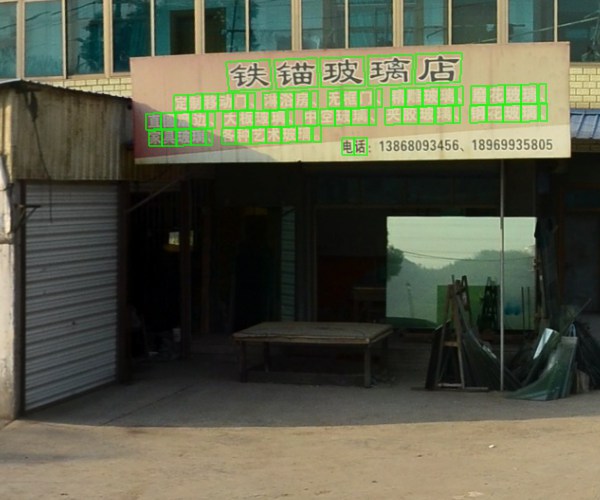} & \hspace{-0.17in}
\includegraphics[width=.11\textwidth]{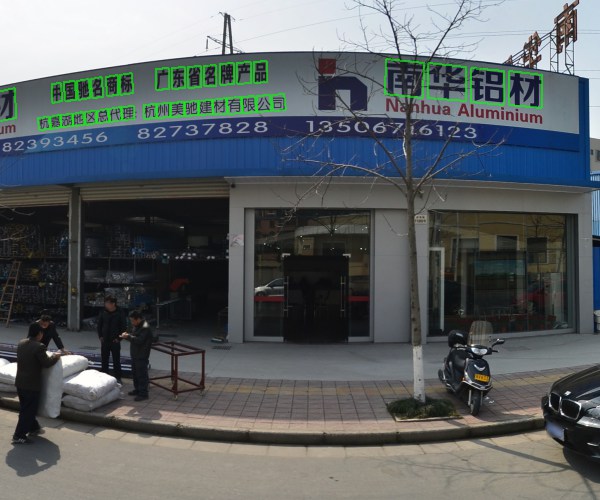} \\
\hspace{-0.1in} \rotatebox{90}{~~~~~~~(b)} & \hspace{-0.17in} \includegraphics[width=.11\textwidth]{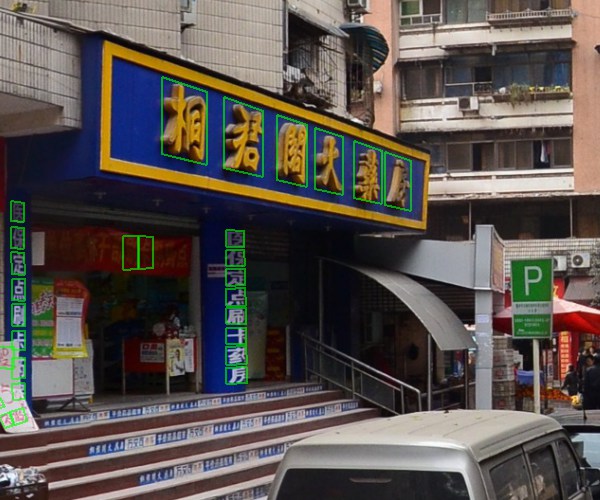} & \hspace{-0.17in}
\includegraphics[width=.11\textwidth]{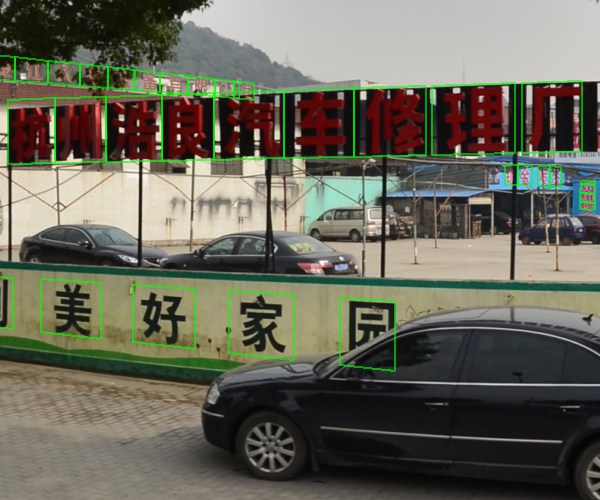} & \hspace{-0.17in}
\includegraphics[width=.11\textwidth]{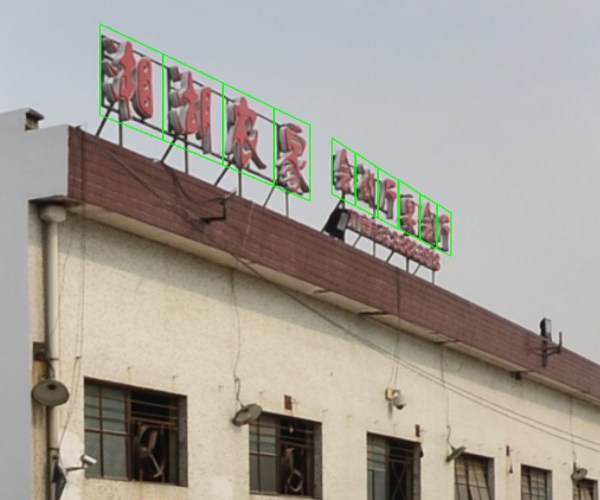} & \hspace{-0.17in}
\includegraphics[width=.11\textwidth]{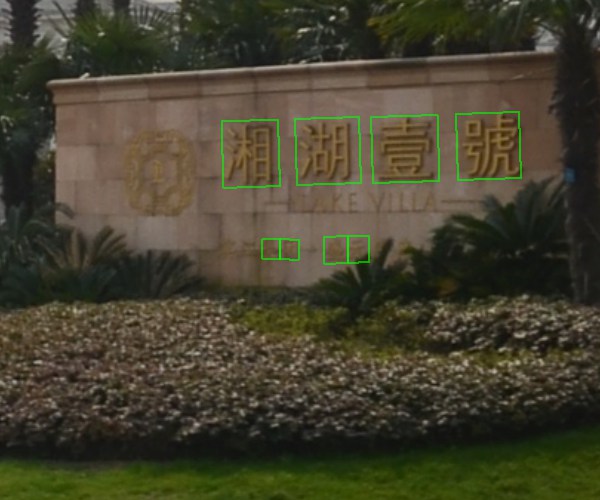} \\
\hspace{-0.1in} \rotatebox{90}{~~~~~~~(c)} & \hspace{-0.17in} \includegraphics[width=.11\textwidth]{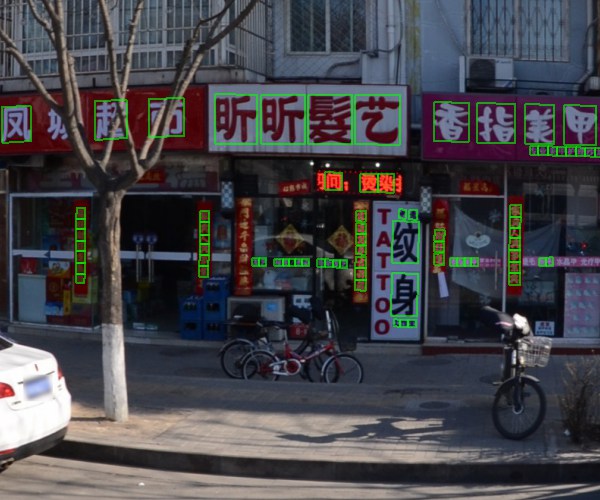} & \hspace{-0.17in}
\includegraphics[width=.11\textwidth]{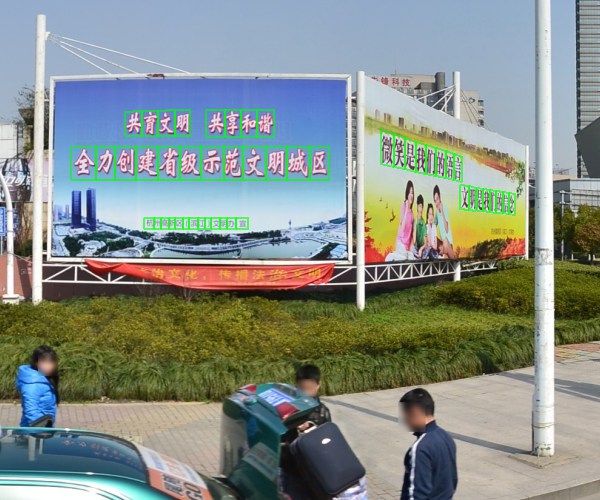} & \hspace{-0.17in}
\includegraphics[width=.11\textwidth]{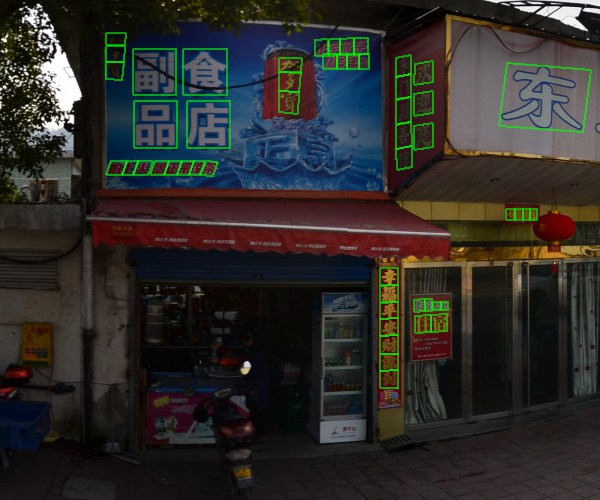} & \hspace{-0.17in}
\includegraphics[width=.11\textwidth]{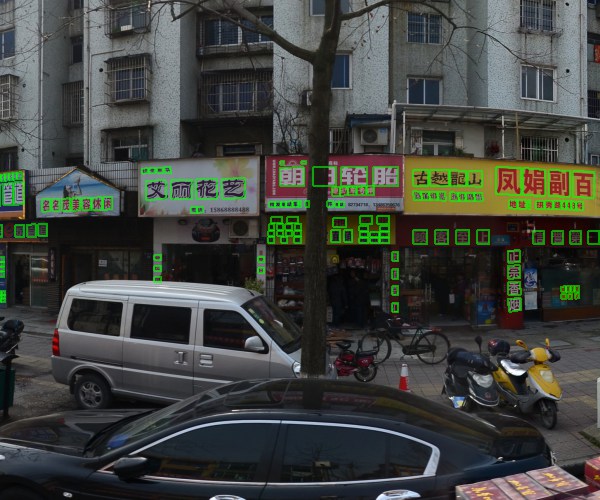}  \\
\hspace{-0.1in} \rotatebox{90}{~~~~~~~(d)} & \hspace{-0.17in} \includegraphics[width=.11\textwidth]{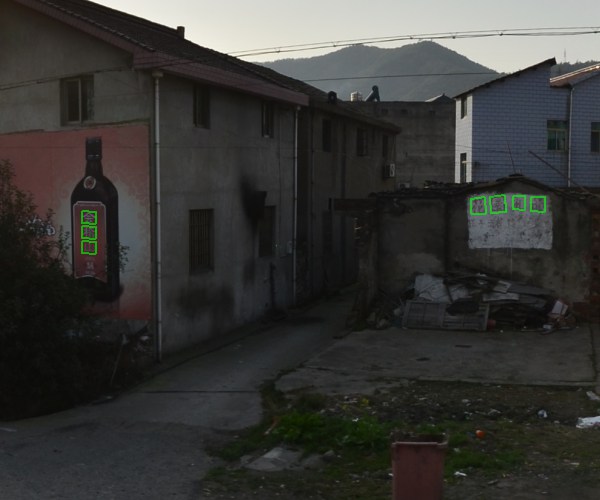} & \hspace{-0.17in}
\includegraphics[width=.11\textwidth]{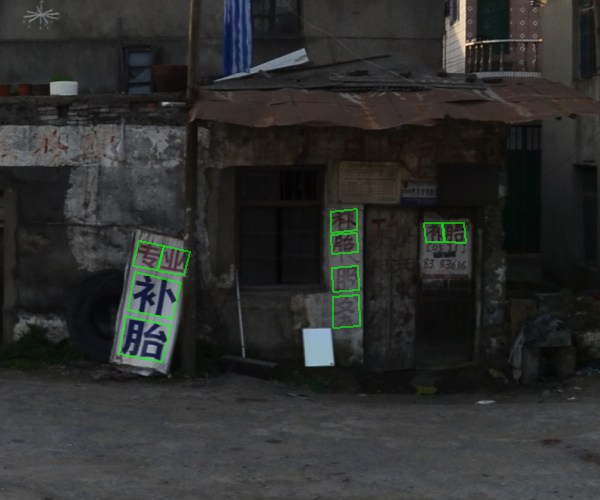} & \hspace{-0.17in}
\includegraphics[width=.11\textwidth]{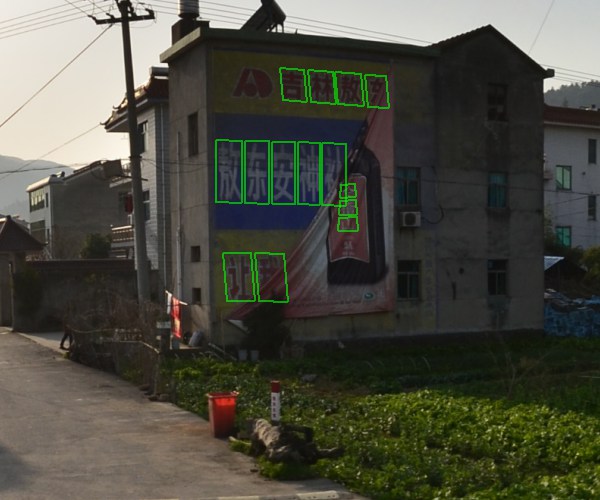} & \hspace{-0.17in}
\includegraphics[width=.11\textwidth]{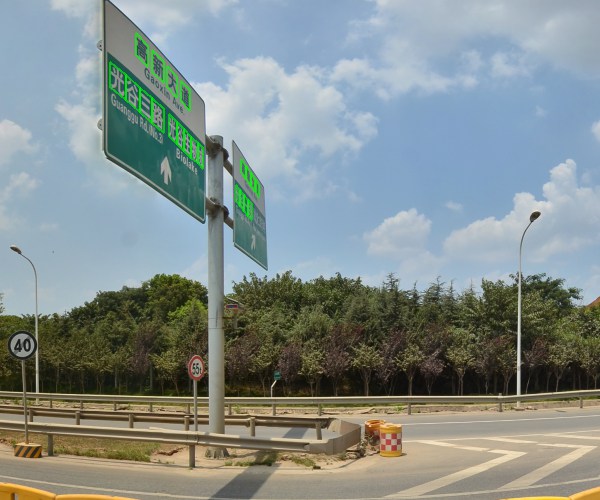}  \\
\hspace{-0.1in} \rotatebox{90}{~~~~~~~(e)} & \hspace{-0.17in} \includegraphics[width=.11\textwidth]{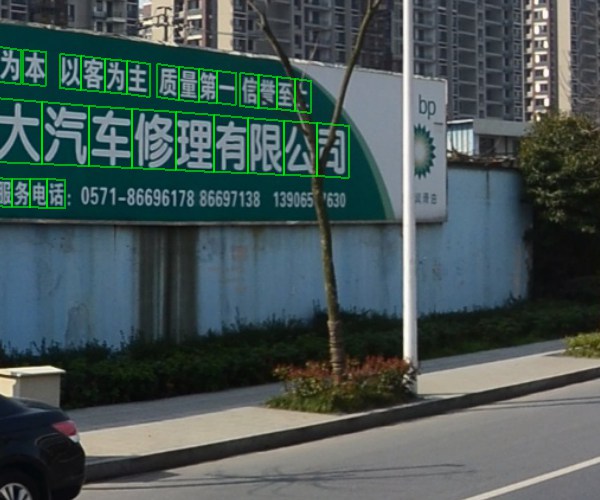} & \hspace{-0.17in}
\includegraphics[width=.11\textwidth]{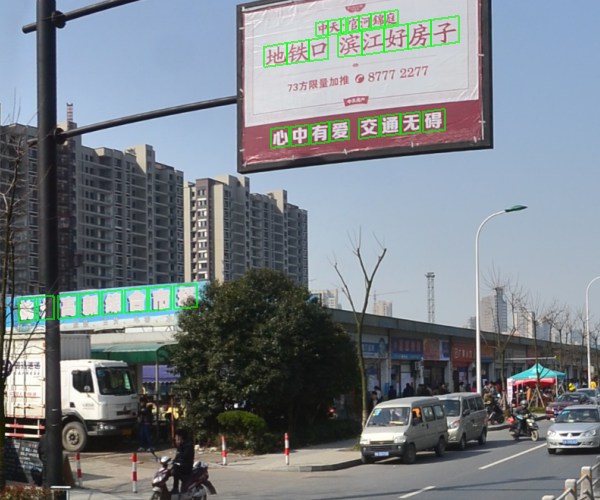} & \hspace{-0.17in}
\includegraphics[width=.11\textwidth]{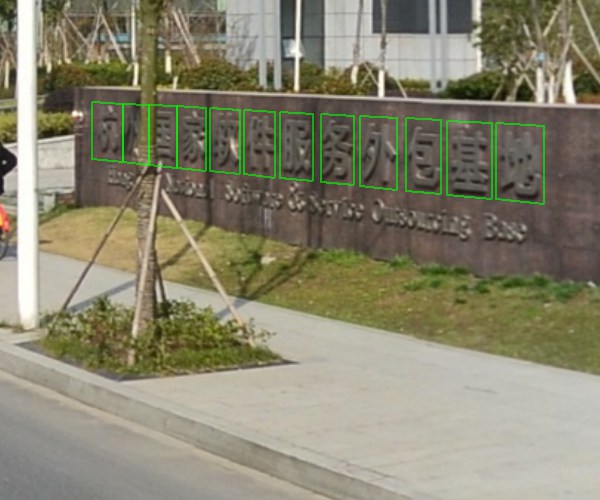} & \hspace{-0.17in}
\includegraphics[width=.11\textwidth]{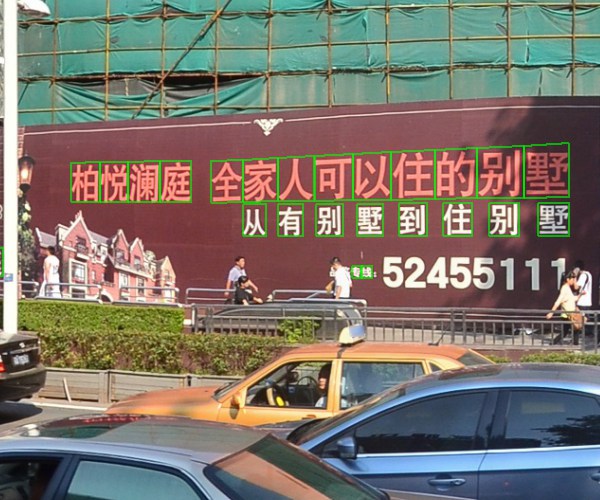}  \\
\hspace{-0.1in} \rotatebox{90}{~~~~~~~(f)} & \hspace{-0.17in} \includegraphics[width=.11\textwidth]{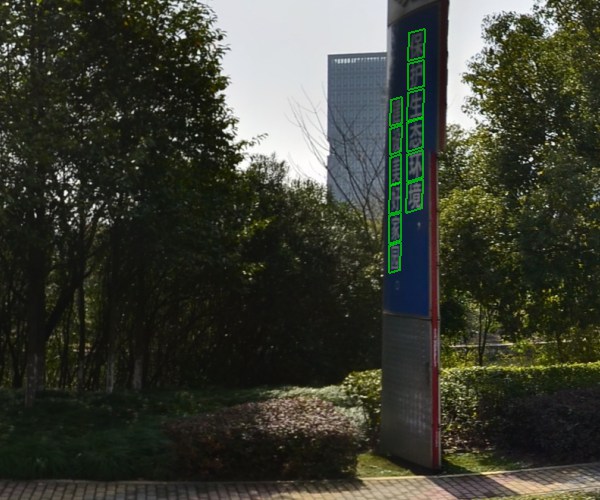} & \hspace{-0.17in}
\includegraphics[width=.11\textwidth]{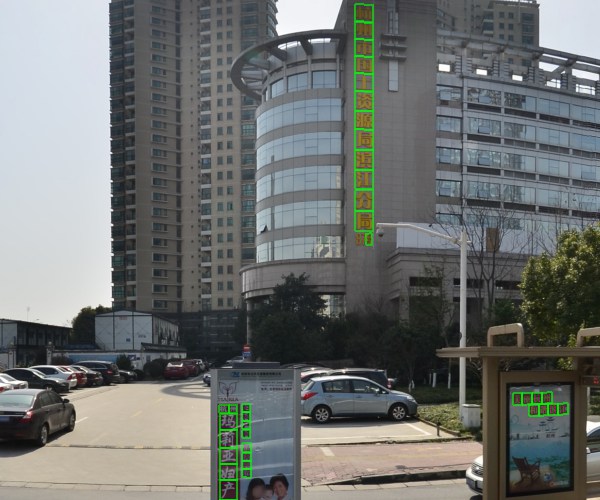} & \hspace{-0.17in}
\includegraphics[width=.11\textwidth]{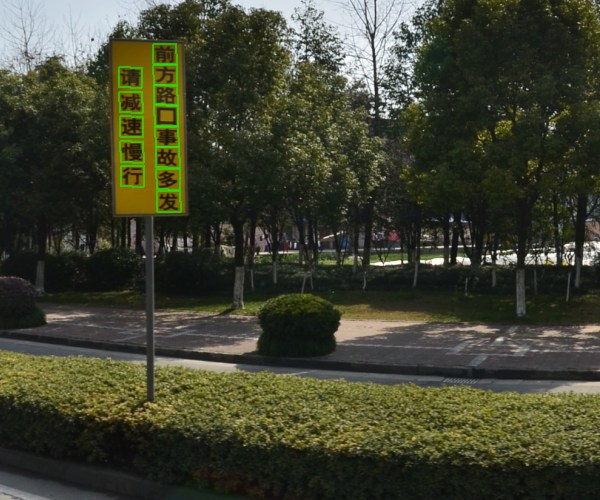} & \hspace{-0.17in}
\includegraphics[width=.11\textwidth]{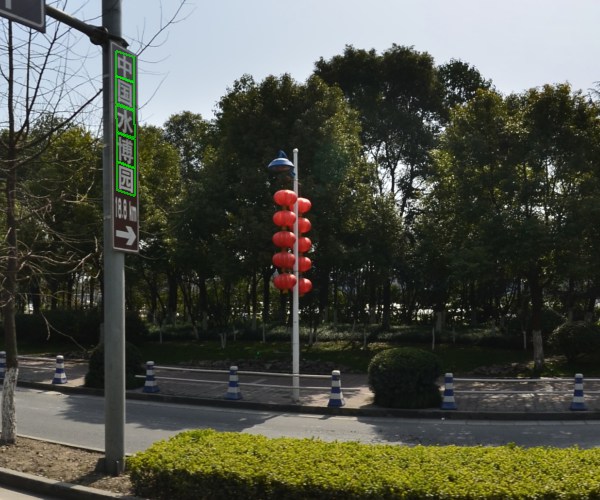}  \\
\hspace{-0.1in} \rotatebox{90}{~~~~~~~(g)} & \hspace{-0.17in} \includegraphics[width=.11\textwidth]{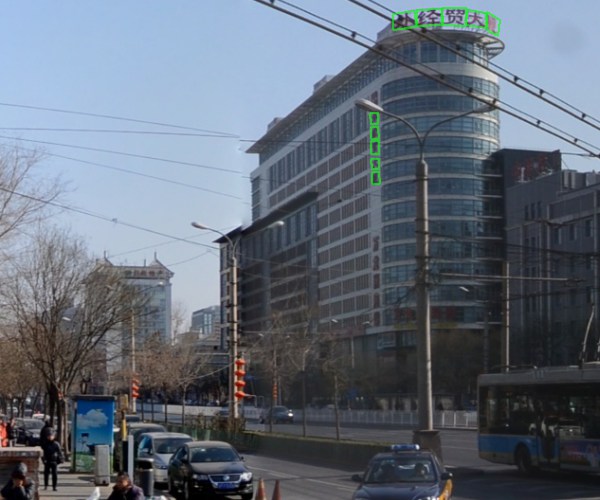} & \hspace{-0.17in}
\includegraphics[width=.11\textwidth]{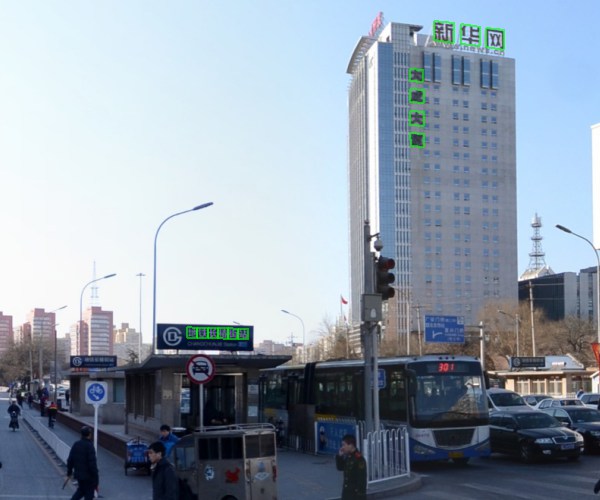} & \hspace{-0.17in}
\includegraphics[width=.11\textwidth]{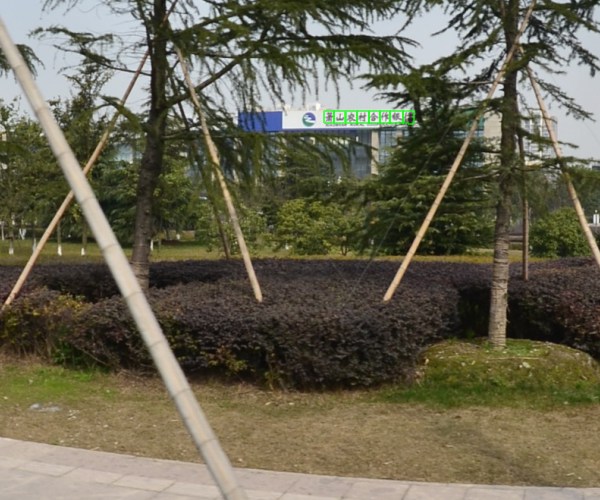} & \hspace{-0.17in}
\includegraphics[width=.11\textwidth]{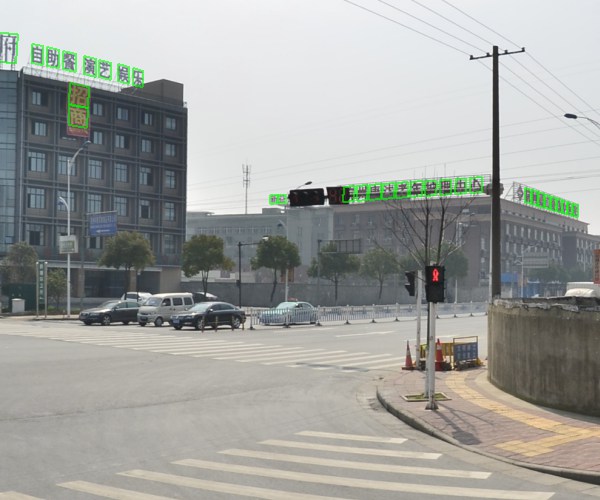} \\
\hspace{-0.1in} \rotatebox{90}{~~~~~~~(h)} & \hspace{-0.17in} \includegraphics[width=.11\textwidth]{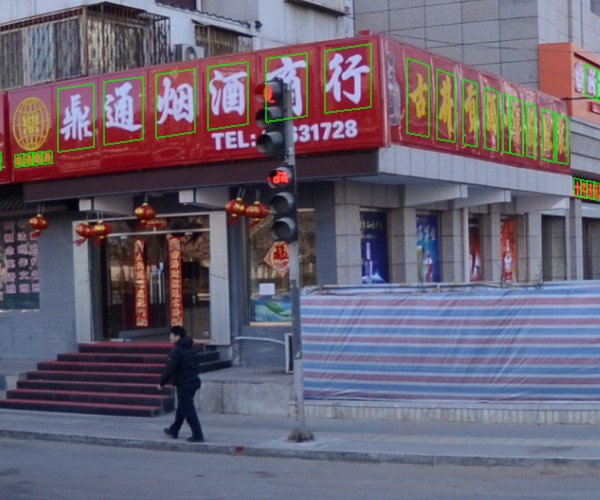} & \hspace{-0.17in}
\includegraphics[width=.11\textwidth]{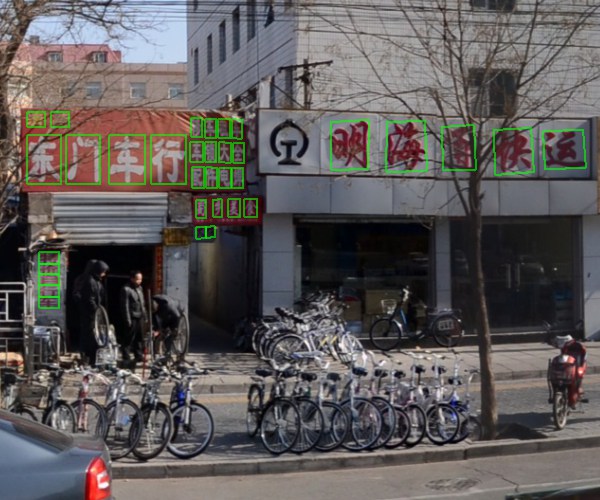} & \hspace{-0.17in}
\includegraphics[width=.11\textwidth]{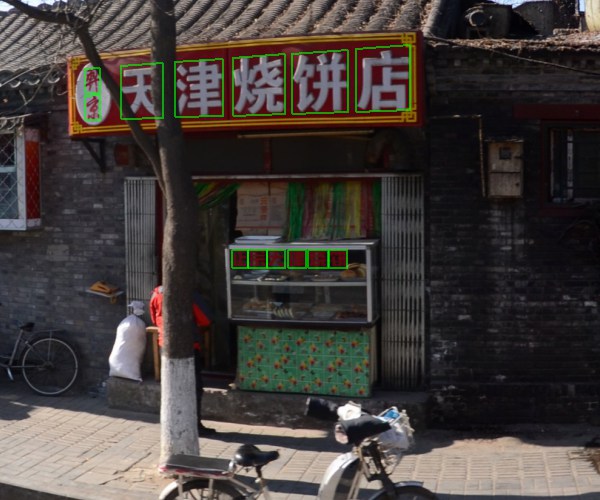} & \hspace{-0.17in}
\includegraphics[width=.11\textwidth]{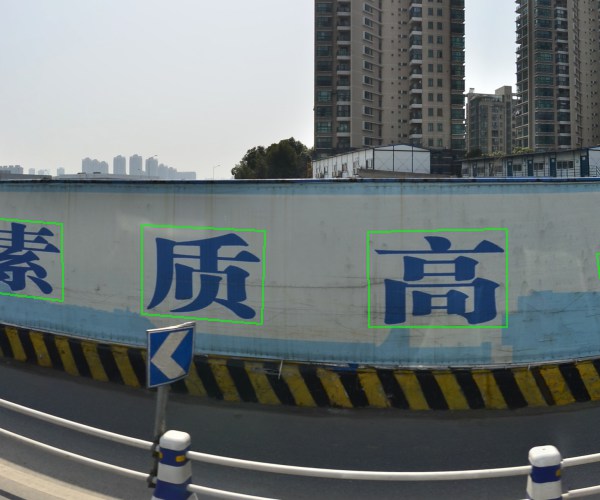} \\
\hspace{-0.1in} \rotatebox{90}{~~~~~~~(i)} & \hspace{-0.17in} \includegraphics[width=.11\textwidth]{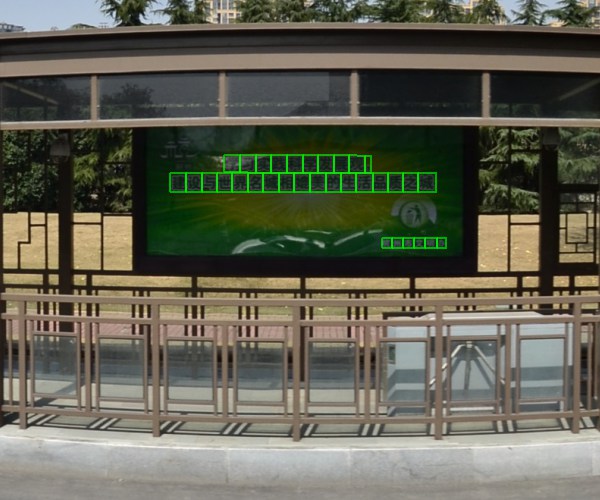} & \hspace{-0.17in}
\includegraphics[width=.11\textwidth]{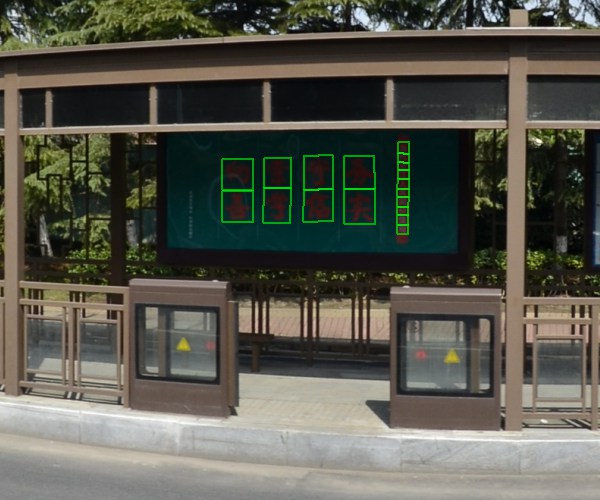} & \hspace{-0.17in}
\includegraphics[width=.11\textwidth]{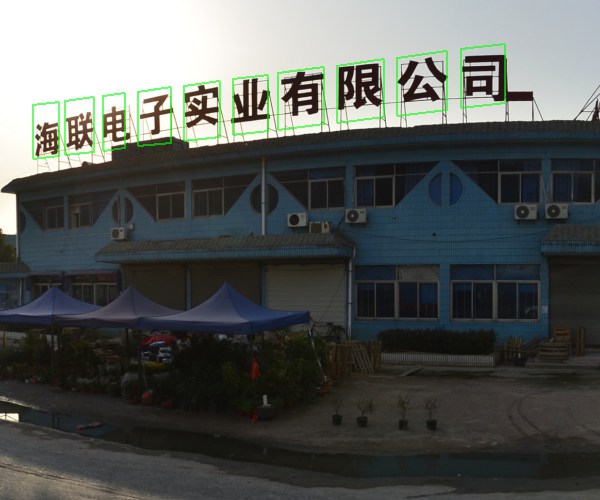} & \hspace{-0.17in}
\includegraphics[width=.11\textwidth]{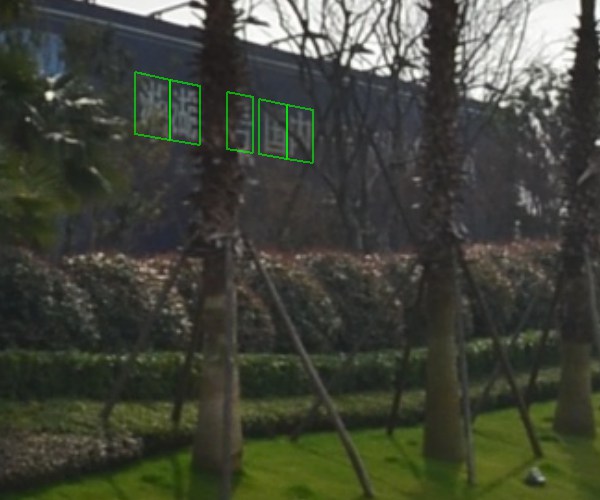} \\
\hspace{-0.1in} \rotatebox{90}{~~~~~~~(j)} & \hspace{-0.17in} \includegraphics[width=.11\textwidth]{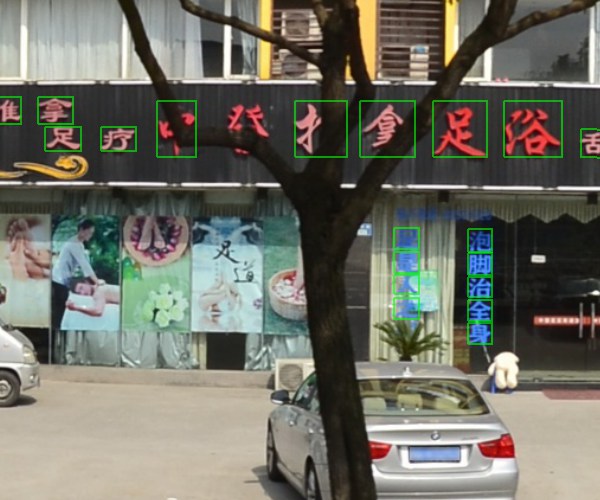} & \hspace{-0.17in}
\includegraphics[width=.11\textwidth]{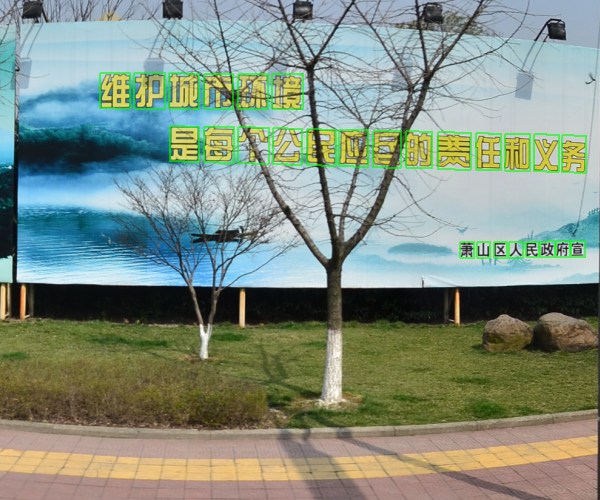} & \hspace{-0.17in}
\includegraphics[width=.11\textwidth]{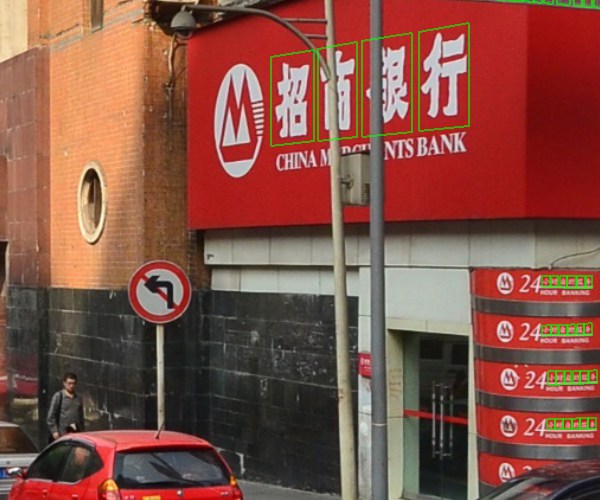} & \hspace{-0.17in}
\includegraphics[width=.11\textwidth]{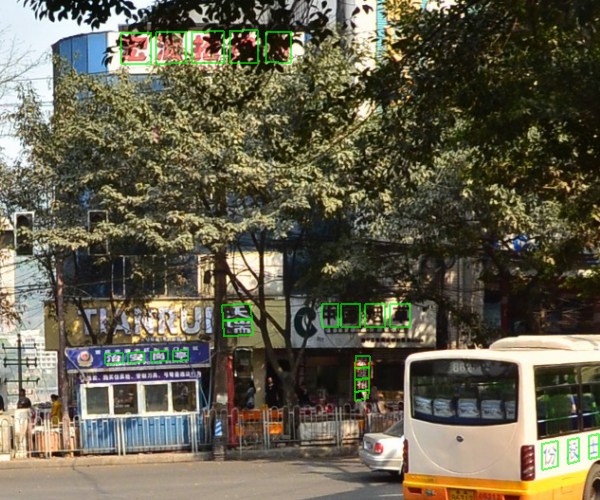}
\end{tabular}
\caption{Dataset diversity. (a) planar text, (b) raised text, (c) text in cities, (d) text in rural areas, (e) horizontal text, (f) vertical text, (g) distant text, (h) nearby text, (i) text under poor illumination, (j) partially occluded text.}
\label{fig:diversity}
\end{figure}

\textbf{Diversity} The above statistics show that our dataset has good diversity on character categories, character sizes, and attributes (i.e., occlusion, background complexity, 3D raised, etc.).
As shown in Figure~\ref{fig:instance_by_attributes}, 13.2\% character instances are occluded, 28.0\% have complex background, 26.0\% are distorted, and 26.9\% are raised text.
As shown in Figure~\ref{fig:diversity}, our dataset contains planar text (a), raised text (b), text in cities (c), text in rural areas (d), horizontal text (e), vertical text (f), distant text (g), nearby text (h), text under poor illumination (i), and partially occluded text (j). Due to such diversity and complexity, our CTW dataset is a challenging dataset.


\section{Baseline Algorithms and Performance} \label{sec:experiment}

We now describe the baseline algorithms and their performance using the proposed CTW dataset.
Our experiments were performed on a desktop with a 3.5GHz Intel Core i7-5930k CPU, NVIDIA GTX TITAN GPU and 32GB RAM.

We considered two tasks: character recognition from cropped regions,  and character detection from images.

\subsection{Recognition} \label{subsec:recognition}

Given a cropped rectangular region showing a Chinese character instance, the goal of the character recognition task is to predict its character category.

\begin{figure}
\centering
\includegraphics[width=.9\linewidth]{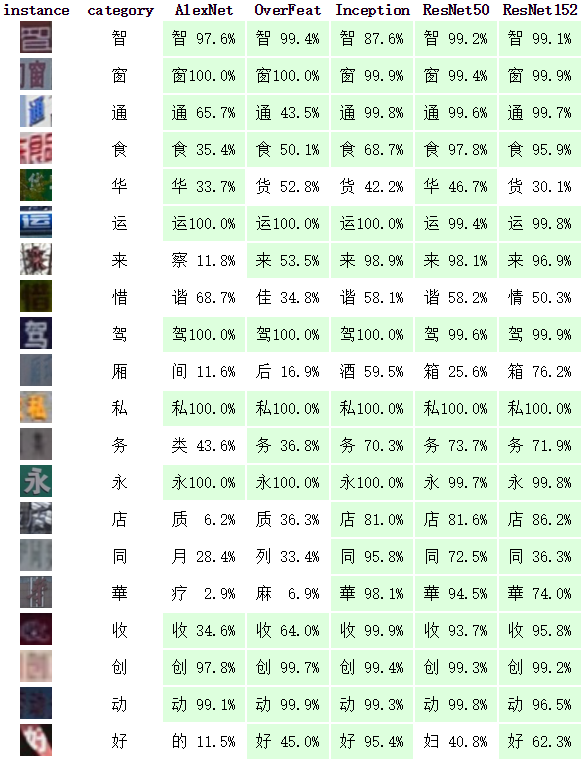}
\caption{Some examples of the recognition task. In each row, from left to right, we give: the cropped region of a character instance, the ground truth character category, and the recognition results of different methods. Corrected recognitions are painted with green. The percentage number shows the confidence of the results.}
\label{fig:result_recognition_task}
\end{figure}

We have tested several state-of-the-art convolutional neural network structures for the recognition task using TensorFlow, including: AlexNet~\cite{NIPS2012_4824}, OverFeat~\cite{DBLP:journals/corr/SermanetEZMFL13}, Google Inception~\cite{DBLP:journals/corr/SzegedyLJSRAEVR14}, 50 layer ResNet~\cite{DBLP:journals/corr/HeZRS15}(ResNet50) and 152 layer ResNet (ResNet152).
We use the training set and the recognition testing set as described in Section~\ref{subsec:data_split} for training and testing, respectively.
Since a majority of the character categories are rarely-used Chinese characters, which have very few samples in the training data and also have very rare usage in practice, we only consider recognition of the top 1000 frequent observed character categories. We consider recognition as a classification problem of 1001 categories. Besides the used 1000 character categories, an 'others' category is added. We trained each network using tens of thousands of iterations, and the parameters of each model are finely tuned. On the testing set, the top-1 accuracy achieved by these networks was: AlexNet (73.0\%), OverFeat (76.0\%), Google Inception (80.5\%), ResNet50 (78.2\%) and ResNet152 (79.0\%), respectively.
In Table~\ref{tab:recognition}, we also give the top-1 accuracy of the top 10 frequent observed character categories. In Figure~\ref{fig:result_recognition_task}, we show 20 character instances randomly chosen from the testing set. In each row, from left to right, we show the cropped region of a character instance, the ground truth character category, and the recognition results of different methods. Among the above methods, Google Inception achieves the best accuracy rate.

\begin{figure*}
    \includegraphics[width=0.95\textwidth]{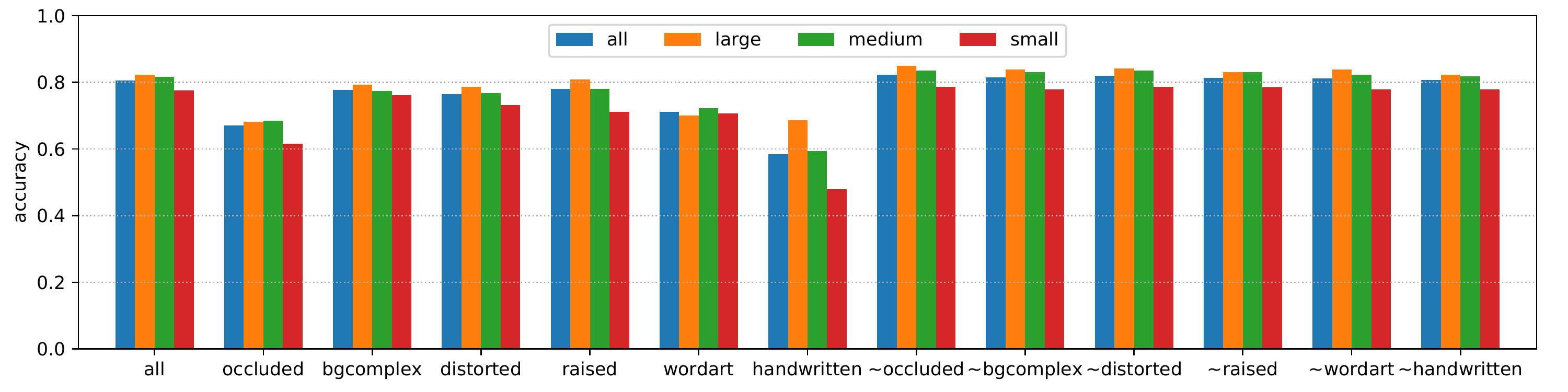}
    \caption{the top-1 accuracy using Google Inception for character instances with different attributes and different sizes. Small, medium, and large refer to character size $<16$, $\in [16,32)$ and $\ge32$, respectively.  $\sim$ denotes without a specific attribute , e.g.,  $\sim$occluded means not occluded character instances.}
    \label{fig:recognition_attributes}
\end{figure*}

In Figure~\ref{fig:recognition_attributes}, we provide the top-1 accuracy using Google Inception for character instances with different attributes and different sizes, respectively.
The results are consistent with our intuition, e.g., characters with clean backgrounds, printed characters, and large characters are easier to recognize than those with complex background, handwritten characters, and small characters, respectively.
An interesting observation is that the recognition accuracy of large wordart characters (70.0\%) is lower than the accuracy of medium wardart characters (72.3\%). The reason is that large characters are more likely to be occluded or have complex background (as shown in Figure~\ref{fig:instance_by_attributes}), making them harder to be recognized.

More details can be found on the website.

\begin{table*}
\centering
\caption{Top-1 accuracy of the 10 most frequent character categories.}
\begin{tabular}{|l|l|l|l|l|l|l|l|l|l|l|}
\hline
          & \begin{minipage}{3.5mm} \includegraphics[width=\linewidth,trim=0 0 0 -2]{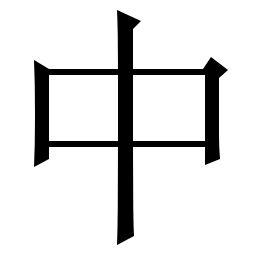} \end{minipage}
          & \begin{minipage}{3.5mm} \includegraphics[width=\linewidth,trim=0 0 0 -2]{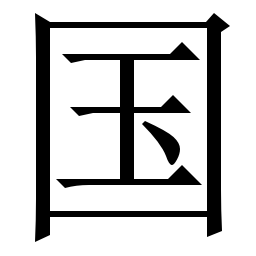} \end{minipage}
          & \begin{minipage}{3.5mm} \includegraphics[width=\linewidth,trim=0 0 0 -2]{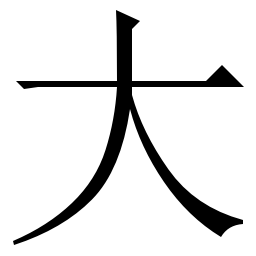} \end{minipage}
          & \begin{minipage}{3.5mm} \includegraphics[width=\linewidth,trim=0 0 0 -2]{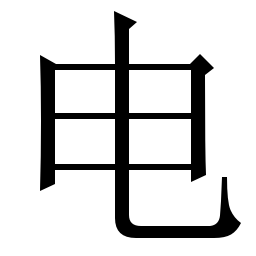} \end{minipage}
          & \begin{minipage}{3.5mm} \includegraphics[width=\linewidth,trim=0 0 0 -2]{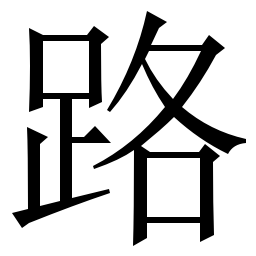} \end{minipage}
          & \begin{minipage}{3.5mm} \includegraphics[width=\linewidth,trim=0 0 0 -2]{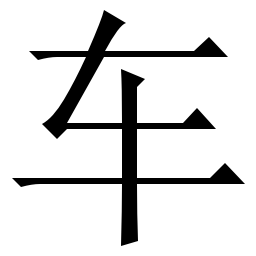} \end{minipage}
          & \begin{minipage}{3.5mm} \includegraphics[width=\linewidth,trim=0 0 0 -2]{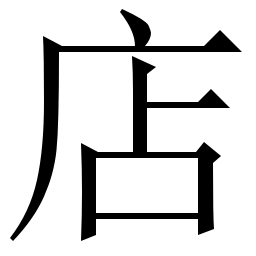} \end{minipage}
          & \begin{minipage}{3.5mm} \includegraphics[width=\linewidth,trim=0 0 0 -2]{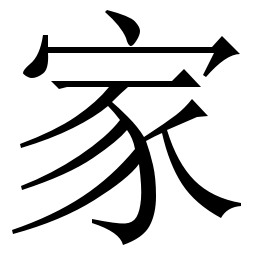} \end{minipage}
          & \begin{minipage}{3.5mm} \includegraphics[width=\linewidth,trim=0 0 0 -2]{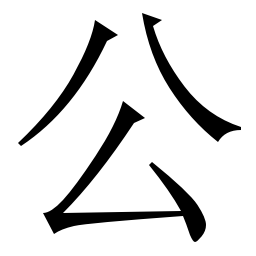} \end{minipage}
          & \begin{minipage}{3.5mm} \includegraphics[width=\linewidth,trim=0 0 0 -2]{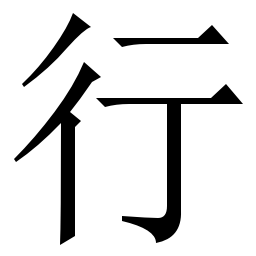} \end{minipage}
           \\ \hline
AlexNet & 79.8\% & 66.2\% & 78.4\% & 83.6\% & 87.3\% & 82.7\% & 79.9\% & 78.9\% & 80.4\% & 84.1\% \\ \hline
OverFeat & 82.7\% & 69.5\% & 84.0\% & 87.2\% & 89.0\% & 86.3\% & 83.4\% & 83.6\% & 82.0\% & 87.1\% \\ \hline
Inception & 88.9\% & 74.6\% & 88.1\% & 90.9\% & 91.2\% & 89.2\% & 90.3\% & 88.4\% & 87.8\% & 90.6\% \\ \hline
ResNet50 & 86.4\% & 72.6\% & 84.0\% & 89.1\% & 90.3\% & 87.1\% & 86.5\% & 84.7\% & 84.1\% & 87.5\% \\ \hline
ResNet152 & 87.4\% & 73.0\% & 85.5\% & 89.3\% & 91.0\% & 87.6\% & 87.1\% & 86.8\% & 84.3\% & 88.4\% \\ \hline
\end{tabular}
\label{tab:recognition}
\end{table*}

\subsection{Detection} \label{subsec:end_to_end_det_and_rec}

Given an image, the goal of the character detection task is to detect the bounding boxes of all character instances and also recognize each character instance, i.e., predict its character category.

We have tested the YOLOv2 algorithm~\cite{DBLP:journals/corr/RedmonF16} for the detection task. Given an image, the output of YOLOv2 is a list of recognized character instances, each is associated with a character category, a bounding box, and a confidence score in $[0,1]$.  We use the training set and the detection testing set as described in Section~\ref{subsec:data_split} for training and testing, respectively. Following the recognition task (Section~\ref{subsec:recognition}), we also limit the number of categories to 1001, i.e., the top 1000 frequent observed character categories and an 'others' category. Since the resolution of images in our dataset is large (i.e., $2048\times 2048$), we have slightly modified YOLOv2 to adapt it to our dataset. For training, first, we set input resolution of YOLOv2 to $672\times 672$. Secondly, each image ($2048\times 2048$ resolution) is uniformly segmented into 196 subimages, each of which has resolution of $168\times 168$ and is overlapped with each other by 23-24 pixels. The subimages are scaled to resolution of $672\times 672$, and then are fed into YOLOv2 as input. For testing, since character instances vary a lot in sizes, in order to detect character instances of different sizes, we perform a multi-scale scheme. First, we set input resolution of YOLOv2 to $1216\times 1216$. Secondly, we segment each input image into 16 subimages ($608\times 608$ resolution) with overlapping of 128 pixels, and also segment the same input image into 64 smaller subimages ($304\times 304$ resolution) with overlapping of 54-55 pixels. After that, all the 80 subimages from both scales are resized to resolution of $1216 \times 1216$ and then are fed into YOLOv2 as input. Finally, non-maximum suppression is applied to remove duplicated detections.

\begin{figure}
\centering
\includegraphics[width=.9\linewidth]{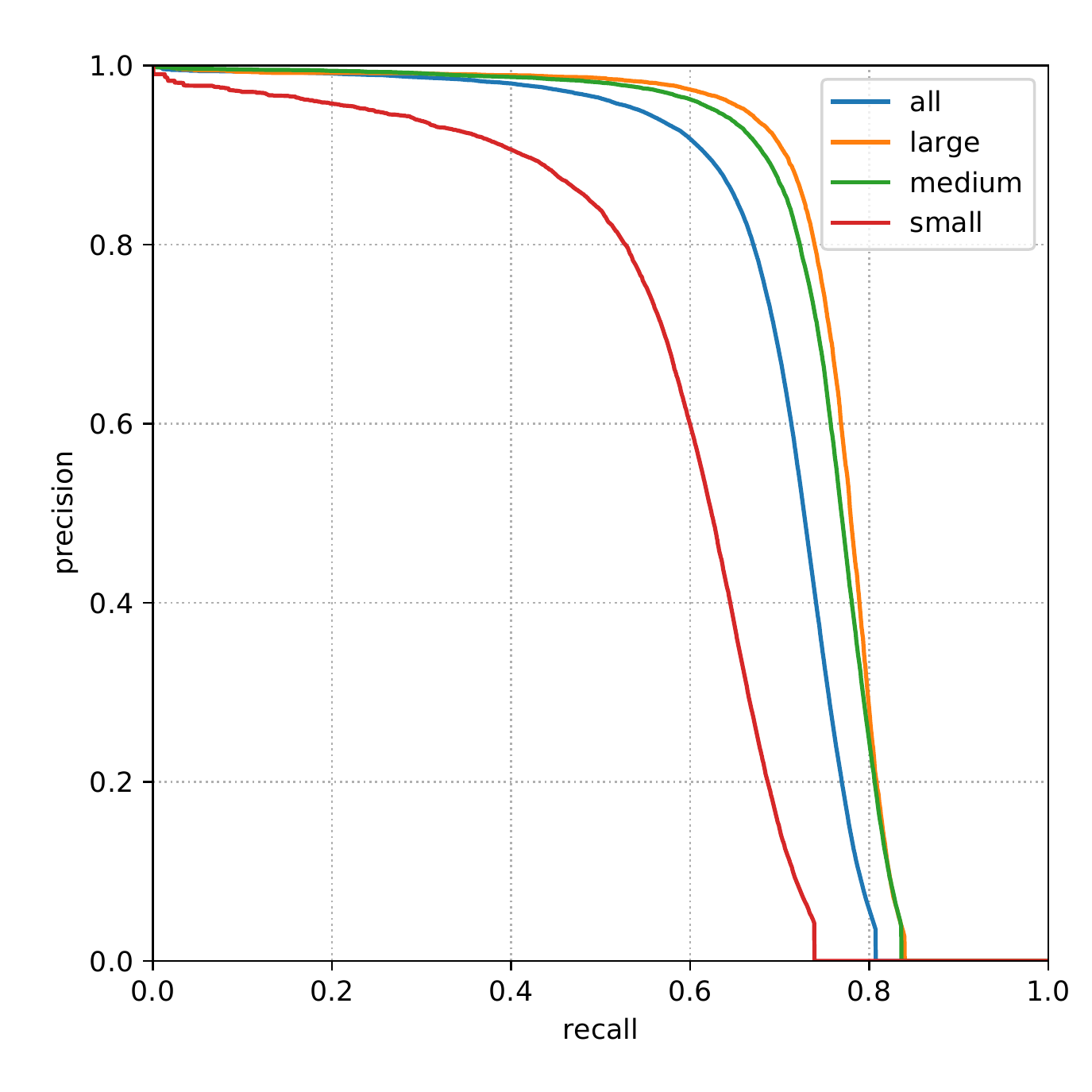}
\caption{Precision-recall curves of the detection task using YOLOv2~\cite{DBLP:journals/corr/RedmonF16}. We show precision-recall curve for all character instances (blue), and curves for character instances with large (yellow), medium (green), and small sizes (red), respectively. }
\label{fig:detection}
\end{figure}

In the testing set, YOLOv2 achieves an mAP of 71.0\%. In Table ~\ref{tab:detection}, we also show the AP scores for the top 10 frequent observed character categories. The AP scores range from 80.3\% to 90.3\%. In Figure~\ref{fig:detection}, we give the overall precision-recall curve, and the precision-recall curve for characters with different sizes.

\begin{figure*}
    \includegraphics[width=0.95\textwidth]{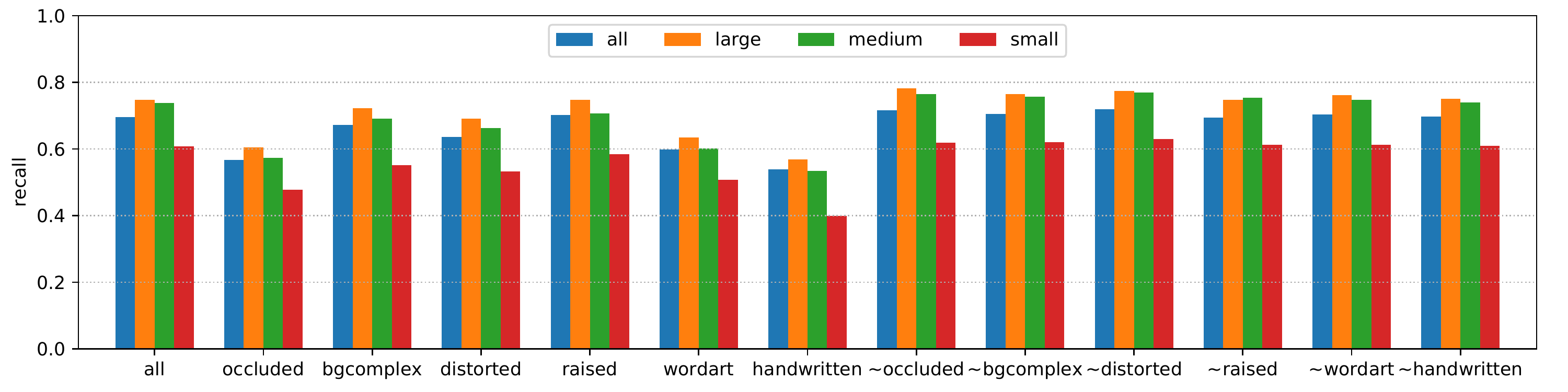}
    \caption{The recall rates of YOLOv2~\cite{DBLP:journals/corr/RedmonF16} for character instances with different attributes and different sizes. Small, medium, and large refer to character size $<16$, $\in [16,32)$ and $\ge32$, respectively.  $\sim$ denotes without a specific attribute , e.g.,  $\sim$occluded means not occluded character instances. }
    \label{fig:detection_attributes}
\end{figure*}

\begin{figure*} [t!]
\centering
\begin{tabular}{c c c c}
\includegraphics[width=.225\linewidth]{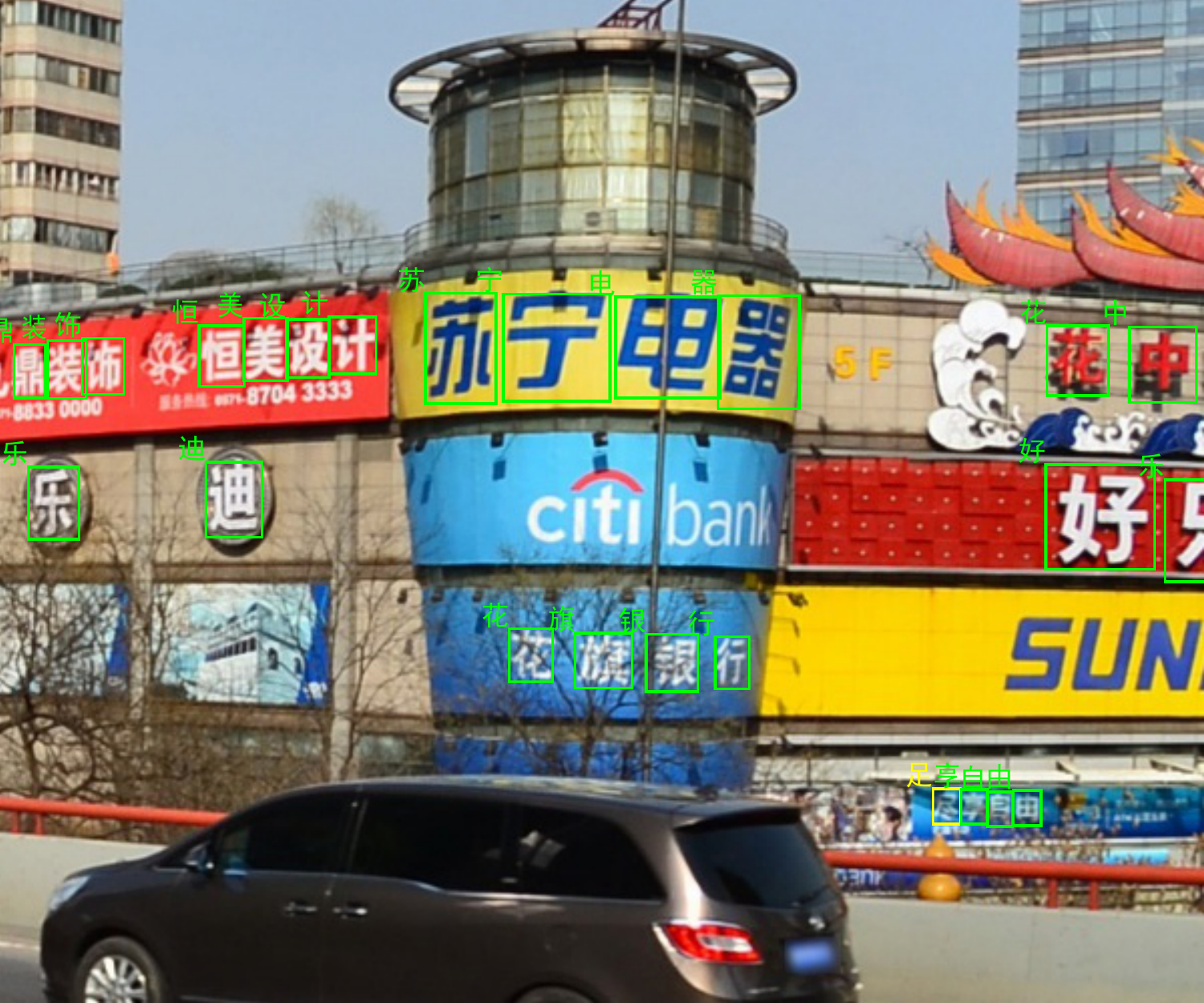} &
\includegraphics[width=.225\linewidth]{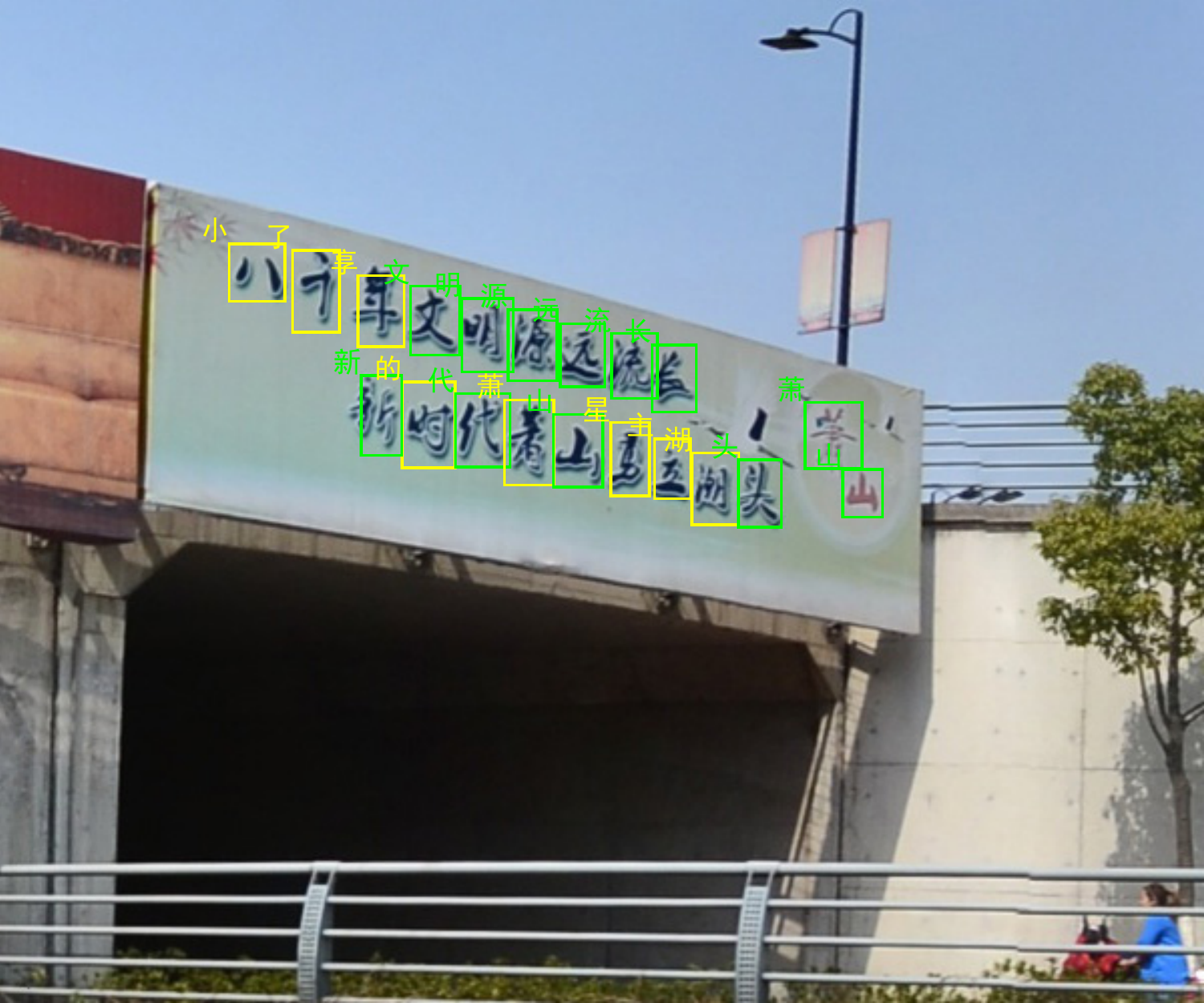}
&
\includegraphics[width=.225\linewidth]{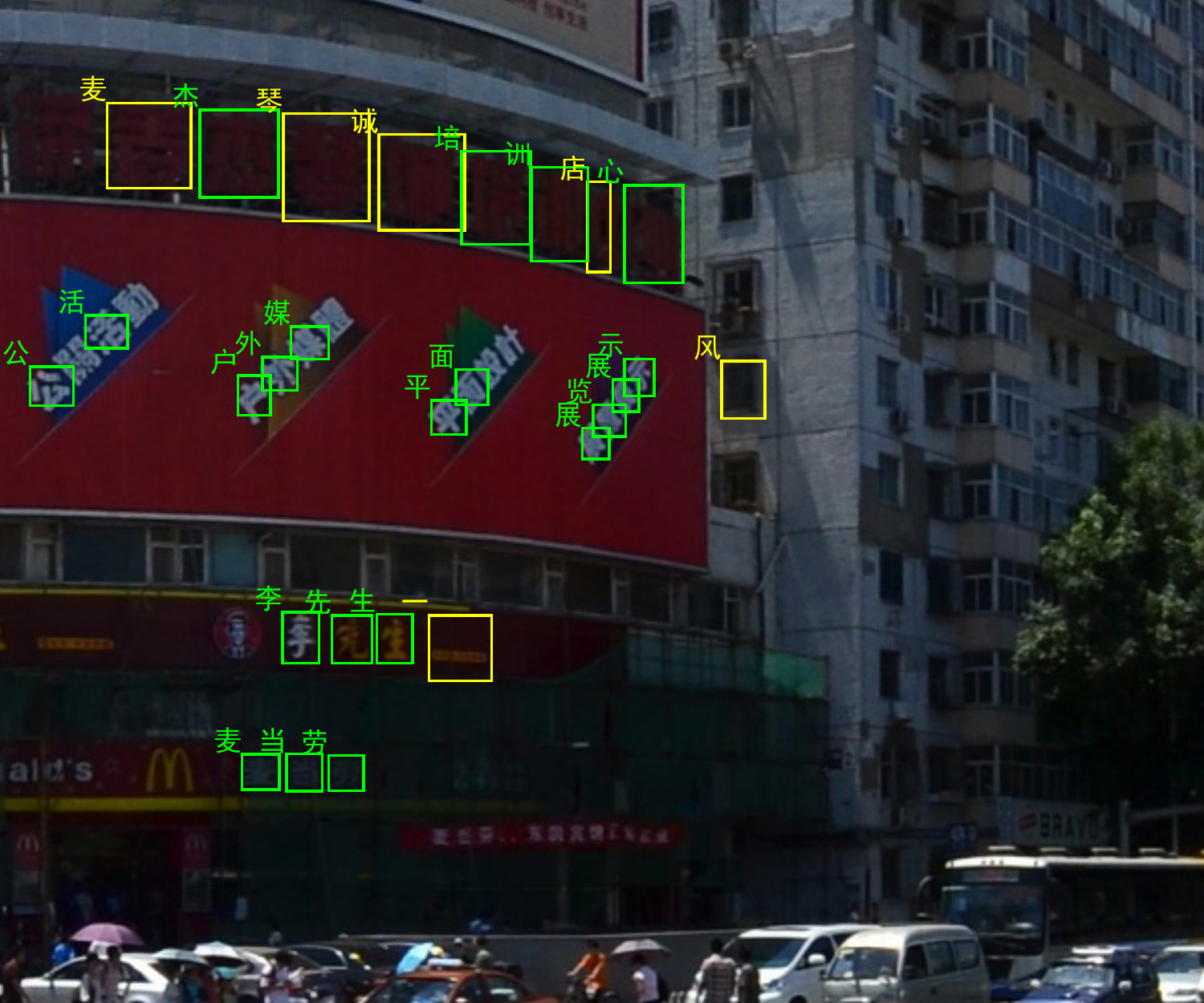} &
\includegraphics[width=.225\linewidth]{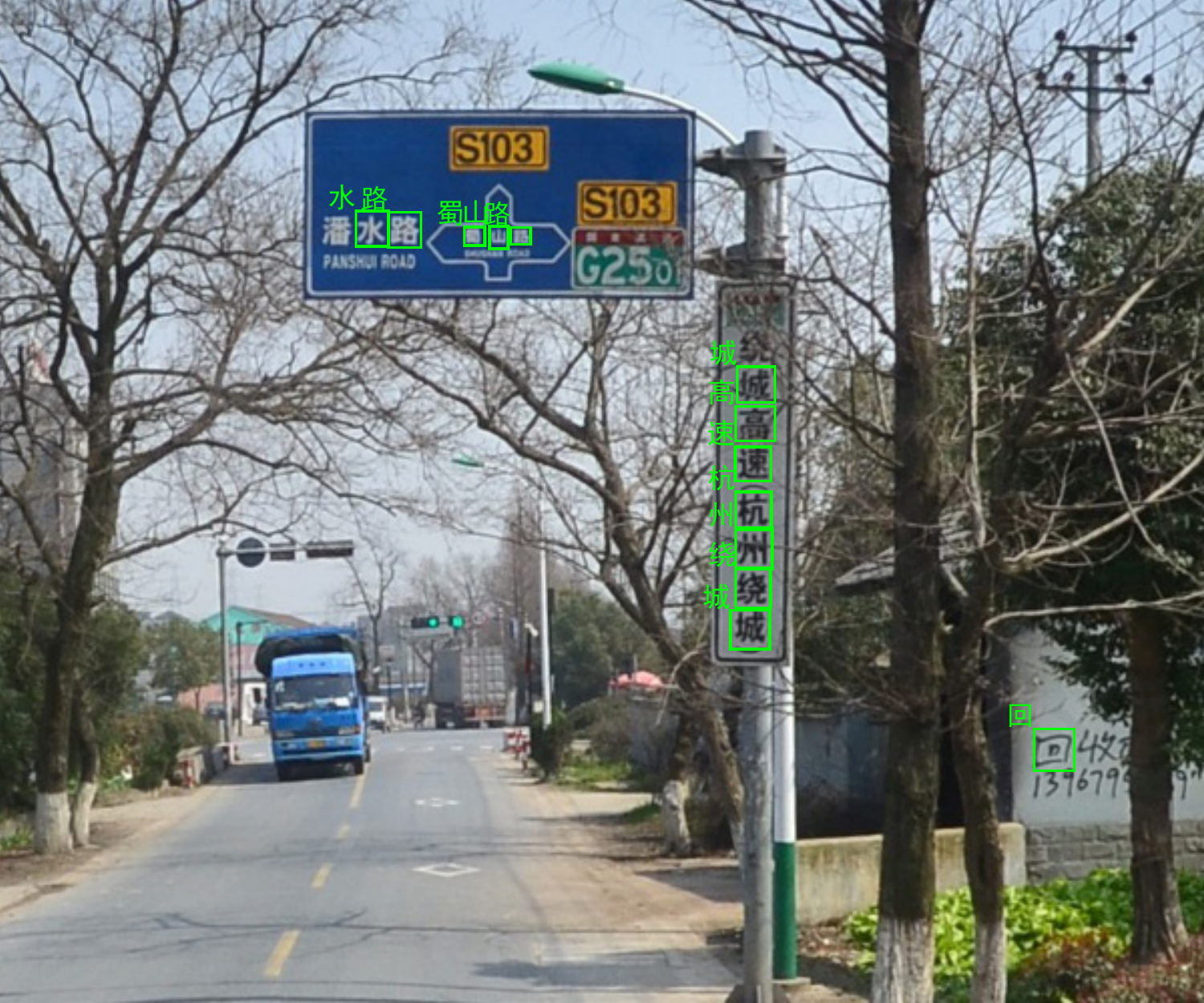}
\\
\includegraphics[width=.225\linewidth]{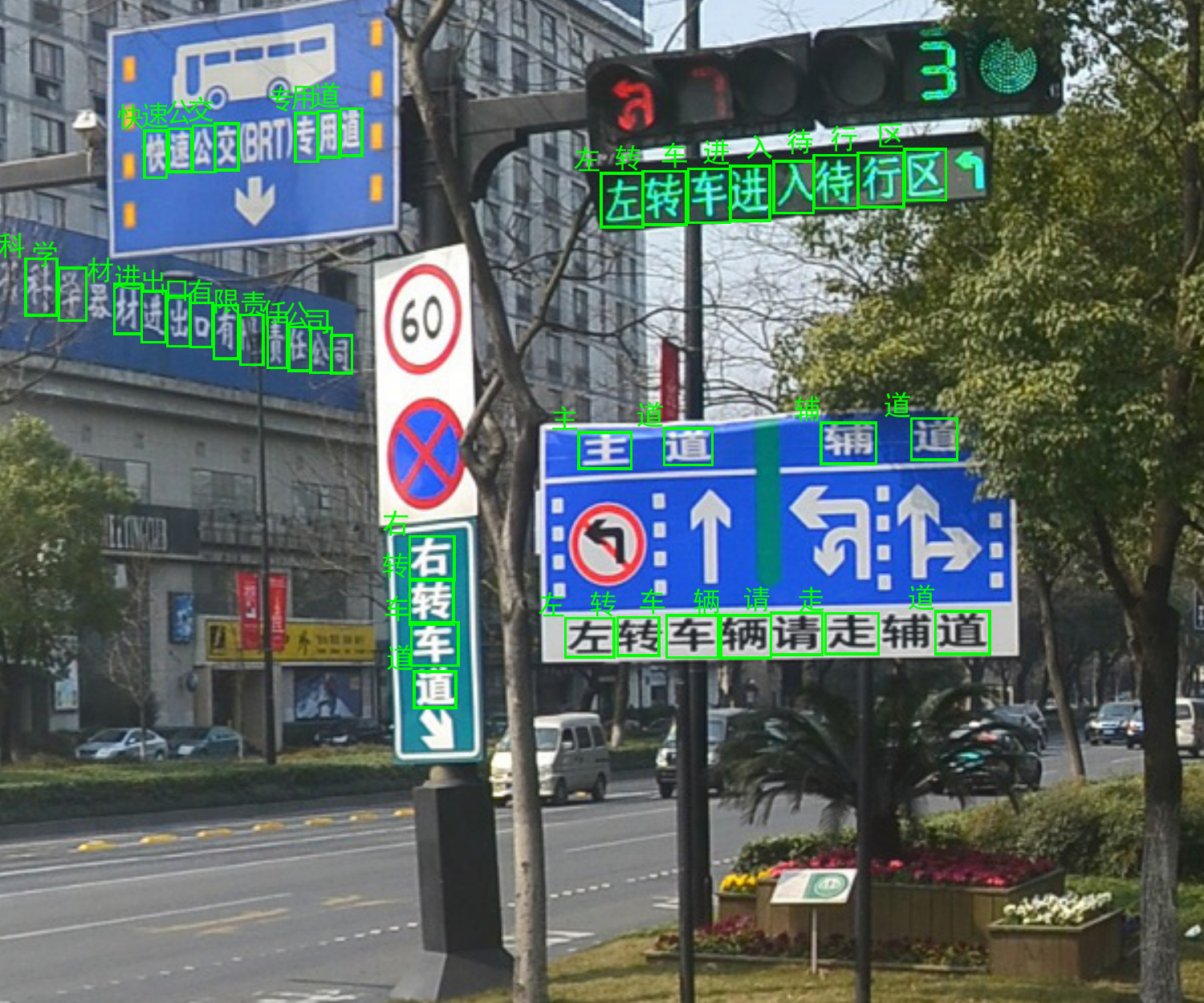} &
\includegraphics[width=.225\linewidth]{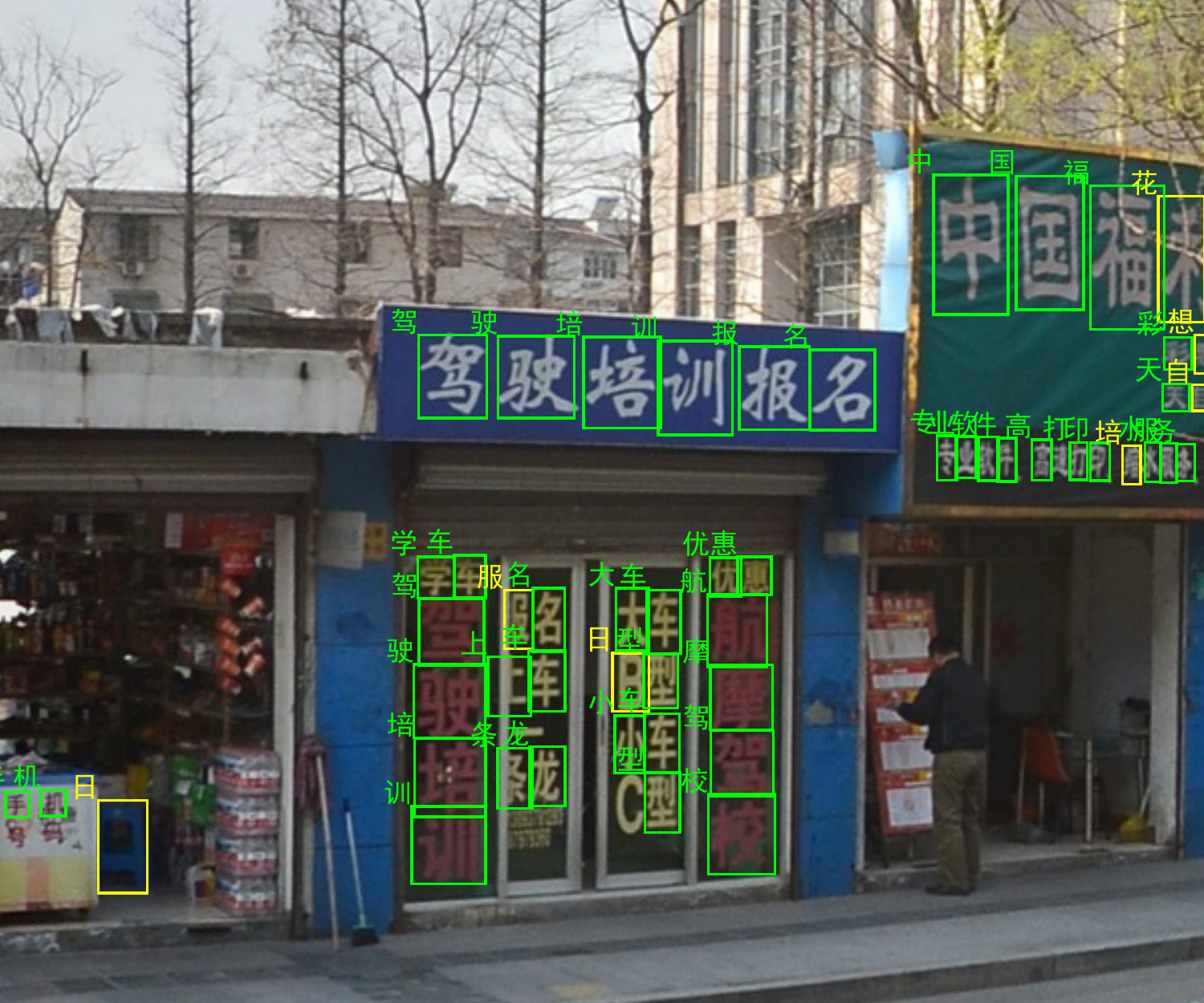}
&
\includegraphics[width=.225\linewidth]{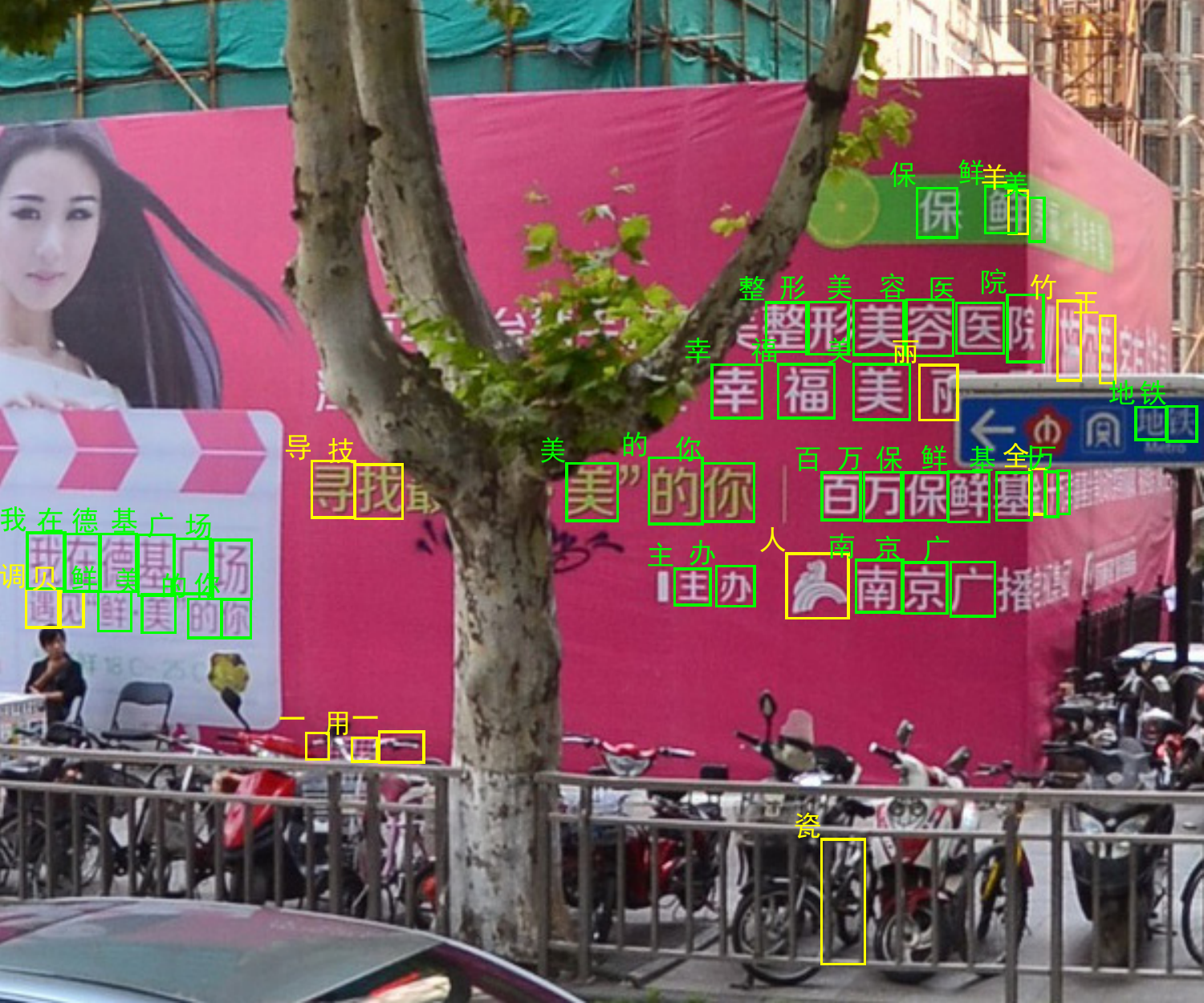} &
\includegraphics[width=.225\linewidth]{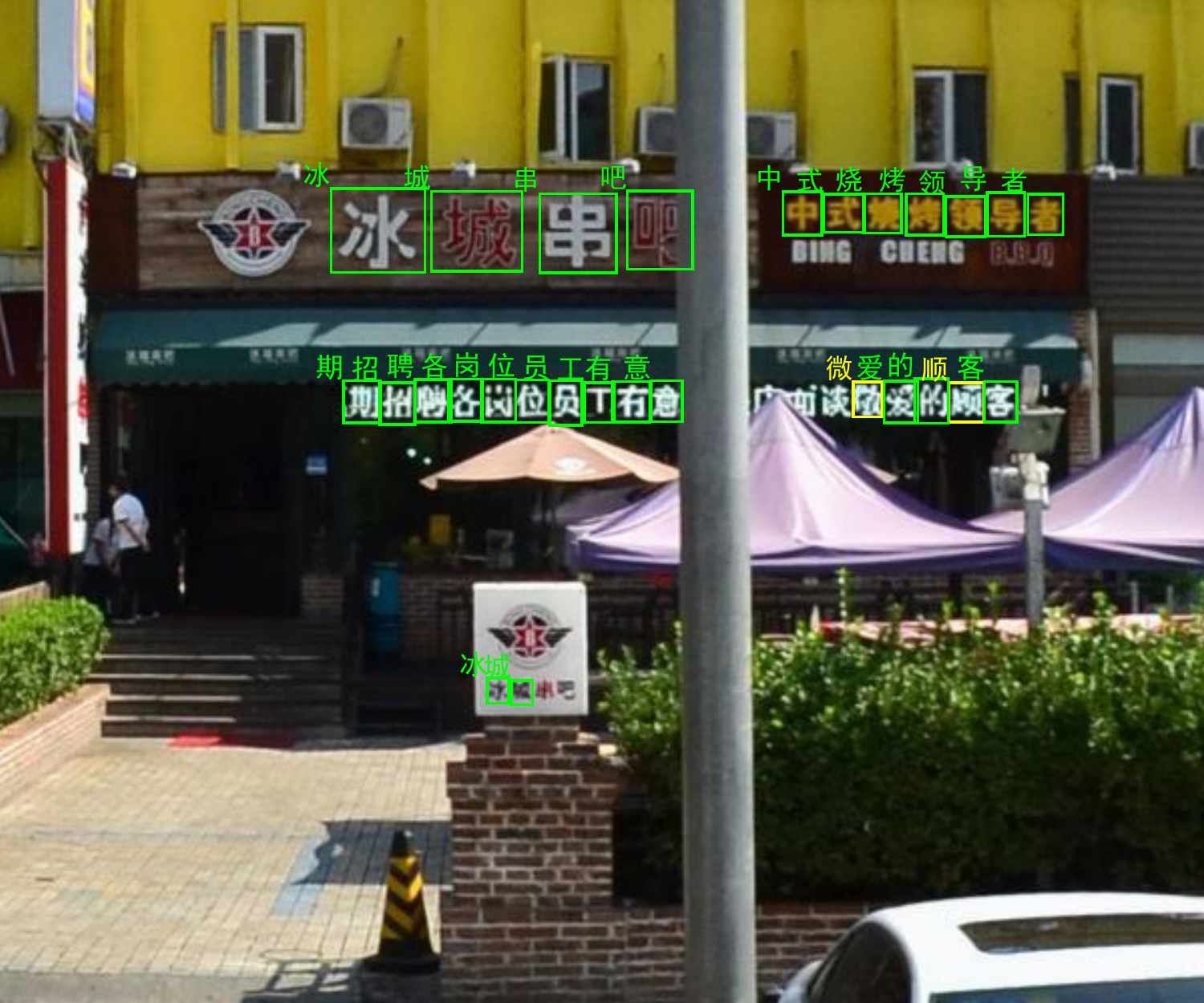}
\end{tabular}
\caption{Detection results by YOLOv2~\cite{DBLP:journals/corr/RedmonF16}. For each image, we give the detected characters and their bounding boxes. Correct detections are shown in green while wrong detections are shown in yellow.
}
\label{fig:result_yolo}
\end{figure*}

In Figure~\ref{fig:detection_attributes}, we provide the recall rates of YOLOv2 for character instances with different attributes and different sizes, respectively. To compute the recall rates, for each image in the testing set, denoting the number of annotated character instances as $n$, we select $n$ recognized character instances with the highest confidences as output of YOLOv2. The results are also consistent with our intuition, i.e., simple characters are easier to be detected and recognized. For example, the recall rates of not occluded characters (71.6\%), printed characters (69.8\%) are higher than the recall rates of occluded characters (56.7\%), handwritten characters (53.8\%), respectively. However, the recall rate of planar characters (69.4\%) is lower than that of raised characters (70.1\%). The reason might be that raised characters have stronger structures than planar characters and hence they are easier to be detected. We also illustrate some detection results of YOLOv2 in Figure~\ref{fig:result_yolo}. More details can be found on the website.

\begin{table*}
\centering
\caption{AP of the 10 most frequent Chinese characters.}
\label{tab:detection}
\begin{tabular}{|l|l|l|l|l|l|l|l|l|l|l|}
\hline
          & \begin{minipage}{3.5mm} \includegraphics[width=\linewidth,trim=0 0 0 -2]{figure/texts/0-0.png} \end{minipage}
          & \begin{minipage}{3.5mm} \includegraphics[width=\linewidth,trim=0 0 0 -2]{figure/texts/0-1.png} \end{minipage}
          & \begin{minipage}{3.5mm} \includegraphics[width=\linewidth,trim=0 0 0 -2]{figure/texts/0-2.png} \end{minipage}
          & \begin{minipage}{3.5mm} \includegraphics[width=\linewidth,trim=0 0 0 -2]{figure/texts/0-3.png} \end{minipage}
          & \begin{minipage}{3.5mm} \includegraphics[width=\linewidth,trim=0 0 0 -2]{figure/texts/0-4.png} \end{minipage}
          & \begin{minipage}{3.5mm} \includegraphics[width=\linewidth,trim=0 0 0 -2]{figure/texts/0-5.png} \end{minipage}
          & \begin{minipage}{3.5mm} \includegraphics[width=\linewidth,trim=0 0 0 -2]{figure/texts/0-6.png} \end{minipage}
          & \begin{minipage}{3.5mm} \includegraphics[width=\linewidth,trim=0 0 0 -2]{figure/texts/0-7.png} \end{minipage}
          & \begin{minipage}{3.5mm} \includegraphics[width=\linewidth,trim=0 0 0 -2]{figure/texts/0-8.png} \end{minipage}
          & \begin{minipage}{3.5mm} \includegraphics[width=\linewidth,trim=0 0 0 -2]{figure/texts/0-9.png} \end{minipage}
          \\ \hline
YOLOv2 & 86.2\% & 81.6\% & 87.0\% & 82.3\% & 89.7\% & 81.6\% & 90.3\% & 81.6\% & 80.3\% & 84.2\% \\ \hline
\end{tabular}
\end{table*}

\section{Conclusions} \label{sec:discussion}

We have introduced Chinese Text in the Wild, a very large dataset of Chinese text in street view images. It contains \totalimage\  images with \totalcharacter\ Chinese character instances, and will be the largest publicly available dataset for Chinese text in natural images. We annotate all Chinese characters in all images. For each Chinese character, the annotation includes its underlying character, the bounding box, and six attributes. We also provide baseline algorithms for two tasks: character recognition from cropped regions, and character detection from images. We believe that our dataset will greatly stimulate future works in Chinese text detection and recognition.

{\small
\bibliographystyle{ieee}
\bibliography{egbib}
}

\end{document}